\documentclass[11pt,letterpaper]{article}

\usepackage{amsmath,amsbsy,amsfonts,amssymb,amsthm,dsfont,fullpage,units}
\usepackage[affil-sl]{authblk}
\usepackage{natbib}
\usepackage{graphicx}
\usepackage{float}
\usepackage{color}
\usepackage{algorithm}
\usepackage{algorithmic}
\usepackage{subfigure} 
\usepackage{multirow}
\usepackage{url}
\usepackage{Definitions}
\usepackage{color}
\usepackage{tikz}
\usetikzlibrary{calc}
\usepackage{psfrag}
\usepackage{exscale,relsize}
\usepackage[normalem]{ulem}
\usepackage{array}

\definecolor{MyBlue}{rgb}{0.69, 0.847, 0.90} 
\newcommand{\R}{\mathbb{R}}
\newenvironment{tight_list}{
\begin{itemize}
\setlength{\itemsep}{1pt}
\setlength{\parskip}{0pt}
\setlength{\parsep}{0pt}
}{\end{itemize}}

\title{Unfolding Latent Tree Structures using 4th Order Tensors}

\author{Mariya Ishteva, Haesun Park, Le Song \\
College of Computing, Georgia Institute of Technology \\
\{mishteva,hpark,lsong\}@cc.gatech.edu
}

\date{}

\begin{document}

\maketitle

\begin{abstract} 
Discovering the latent structure from many observed variables is an important yet
challenging learning task. 
Existing approaches for discovering latent structures
often require the unknown number of hidden states as an input.
In this paper, we propose a quartet based approach which is \emph{agnostic} to this number. 
The key contribution is a novel rank characterization of the tensor associated with the marginal distribution of a quartet. This characterization allows us to design a \emph{nuclear norm} based test for resolving quartet relations. We then 
use the quartet test as a subroutine in a divide-and-conquer algorithm for recovering the latent tree structure. Under mild conditions, the algorithm is consistent and its error probability decays exponentially with increasing sample size. We demonstrate that the proposed approach compares favorably to alternatives. In a real world stock dataset, it also discovers meaningful groupings of variables, and produces a model that fits the data better.
\end{abstract} 

\section{Introduction}
\label{sec:introduction}

Discovering the latent structure from many observed variables is an important yet challenging learning task. The discovered structures can help better understand the domain and lead to potentially better predictive models. Many local search heuristics based on maximum parsimony and maximum likelihood methods have been proposed to address this problem~\citep{SemSte03,Zhang04,HelGha05,TehDauRoy08,HarWil10}. Their common drawback is that it is difficult to provide consistency guarantees. Furthermore, the number of hidden states often needs to be determined before the structure learning. Or cross-validations are needed to determine the hidden states, which can be very time consuming to run.

Efficient algorithms with provable performance guarantees have been explored in the phylogenetic tree reconstruction community. One popular algorithm is the neighbor-joining (NJ) algorithm~\citep{SaiNei87}, where pairs of variables are joined recursively according to a certain distance measure. The NJ algorithm is consistent when the distance measure satisfies the path additive property~\citep{MihLevPac2009}. For discrete random variables, the additive distance is defined using the determinant of the joint probability table of a pair of variables~\citep{Lake1994}. However, this definition only applies to the cases where the observed variables and latent variables have the same number of states. When the latent variables represent simpler factors with smaller number of states, the NJ algorithm can perform poorly. 

Another family of provably consistent reconstruction methods is the quartet-based methods \citep{SemSte03,ErdSzeSteWar99b}. These methods first resolve a set of latent relations for quadruples of observed variables (quartets), and subsequently, stitch them together to form a latent tree. A good quartet test plays an essential role in these methods, as it is called repeatedly by the stitching algorithms. Recently,~\citep{AnaChaHsuKakSonZha2011} proposed a quartet test using the leading $k$ singular values of the joint probability table, where $k$ is the number of hidden states. This new approach allows $k$ to be different from the number of the observed states. However, it still requires $k$ to be given in advance. 

Our goal is to design a latent structure discovery algorithm which is \emph{agnostic} to the number of hidden states, since in practice we rarely know this number. The proposed approach is quartet based, where the quartet relations are resolved based on rank properties of $4$th order tensors associated with the joint probability tables of quartets. The key insight is that rank properties of the tensor reveal the latent structure behind a quartet. Similar observations have been reported in the phylogenetic community~\citep{E05,AllRho06}, but they are concerned about the cases where the number of hidden states is larger or equal to the number of observed states. 
We focus instead on the cases where the number of hidden states is smaller, representing simpler factors. Furthermore, if the joint probability tensor is only approximately given (due to sampling noise) the main rank condition has to be modified. In~\citet{AllRho06} such condition is missing and in~\citet{E05} the condition is heuristically translated to the distance of a matrix to its best rank-$k$ approximation. In contrast, we propose a novel nuclear norm relaxation of the rank condition, discuss its advantages, and provide recovery conditions and finite sample guarantees. 
Our quartet test is easy to compute since it only involves singular value decomposition of unfolded $4$th order tensors. 

Using the proposed quartet test as a subroutine, the latent tree structure can be recovered in a divide-and-conquer fashion~\citep{PeaTar86}. For $d$ observed variables, the computational complexity of the algorithm is $O(d \log d)$, making it scalable to large problems. 
Under mild conditions, the tree construction algorithm using our quartet test is consistent and stable to estimate given a finite number of samples. In simulations, we compared to alternatives in terms of resolving quartet relations and building the entire latent trees. The proposed approach is among the best performing ones while being agnostic to the number of hidden states $k$. The latter is an important improvement, since cross validation for finding $k$ is expensive while leading to similar final results. We also applied the new approach to a stock dataset, where it discovered meaningful grouping of stocks according to industrial sectors, and led a latent variable model that fits the data better than the competitors. 

\section{Latent Tree Graphical Models}
\label{sec:latent_tree}

In this paper, we focus on discrete latent variable models where the conditional independence structures are specified by trees. 
We assume that the $d$ observed variables, $\Oscr = \cbr{X_1,\ldots,X_d}$, are leaves of the tree and that they all have the same number of states, $n$. We also assume the  $d_h$ hidden variables, $\Hscr = \cbr{X_{d+1},\ldots,X_{d+d_h}}$, 
have the same\footnote{Our results are easily generalizable to the case where all hidden variables have different number of states.}, 
\emph{but unknown}, number of states, $k$, ($k\leq n$). Furthermore, we use uppercase letters to denote random variables (\eg, $X_i$) and lowercase letters their instantiations (\eg, $x_i$). 

{\bf Factorization of distribution.} The joint distribution of all variables, $\Xscr=\Oscr \cup \Hscr$, in a latent tree model is a multi-way table (tensor), $\Pcal$, with $d+d_h$ dimensions. Although the tensor has $O(n^d k^{d_h})$ number of entries, they can be computed from just a polynomial number of parameters due to the latent tree structure. That is $\Pcal(x_1,\ldots,x_{d+d_h}) = \prod_{i=1}^{d+d_h} P(x_i | x_{\pi_i})$ where each $P(X_i|X_{\pi_i})$ is a conditional probability table (CPT) of a variable $X_i$ and its parent $X_{\pi_i}$ in the tree.\footnote{For a latent tree, we can select a latent node as the root, and re-orient all edges away from it to induce consistent parent-child relations. For the root node $X_r$, $P(X_r | X_{\pi_r})=P(X_r)$.} This factorization leads to a significant saving in terms of tensor representation: we can represent exponential number of entries using just $O(d_h k^2 + dnk)$ parameters from the CPTs. Throughout the paper, we assume that {\bf (A1)} all CPTs have full column rank, $k$. 

{\bf Structure learning.} Determining the tree topology $\Tcal$ is an important and challenging learning problem. The goal is to discover the latent structure based just on samples from observed variables. For simplicity and uniqueness of the tree topology~\citep{Pearl88}, we assume that {\bf (A2)} every latent variable has \emph{exactly} 3 neighbors.

\begin{figure}[ht]
\centering
    \renewcommand{\arraystretch}{1}     
    \setlength{\tabcolsep}{5pt}  
    \begin{tikzpicture}
    [
      scale=0.75,
      observed/.style={circle,inner sep=0.01mm,draw=black,fill=MyBlue}, 
      hidden/.style={circle,inner sep=0.01mm,draw=black},
      hidden2/.style={circle,inner sep=0.3mm,draw=black},
      hidden3/.style={circle,inner sep=1.2mm,draw=black},
    ]
    \node [observed,name=z1] at (-3.1,0.8) {$\mathsmaller X_{i_1}$};
    \node [observed,name=z2] at (-3.1,-0.8) {$\mathsmaller X_{i_2}$};
    \node [observed,name=z3] at (1.4,0.8) {$\mathsmaller X_{i_3}$};
    \node [observed,name=z4] at (1.4,-0.8) {$\mathsmaller X_{i_4}$};
    \node [hidden,name=h] at ($(-1.7,0)$) {$\mathsmaller H_i$};
    \node [hidden3,name=m] at ($(-1,0)$){$ $}; 
    \node [hidden,name=g] at ($(0,0)$) {$\mathsmaller G_i$};
    \node [hidden3,name=empty1] at ($(-2.4,0.4)$) {$ $};
    \node [hidden3,name=empty2] at ($(-2.4,-0.4)$) {$ $};
    \node [hidden3,name=empty3] at ($(0.7,0.4)$) {$ $};
    \node [hidden3,name=empty4] at ($(0.7,-0.4)$) {$ $};
    \draw [-] (z1) to (empty1);
    \draw [line width=0.4mm,style=dotted] (empty1) to (h);
    \draw [-] (z2) to (empty2);
    \draw [line width=0.4mm,style=dotted] (empty2) to (h);
    \draw [-] (z3) to (empty3);
    \draw [line width=0.4mm,style=dotted] (empty3) to (g);
    \draw [-] (z4) to (empty4);
    \draw [line width=0.4mm,style=dotted] (empty4) to (g);
    \draw [-] (h) to (m);
    \draw [line width=0.4mm,style=dotted]  (m) to (g);
    \end{tikzpicture}
    \caption{Quartet ($X_1$, $X_2$, $X_3$, $X_4$) from a tree.}
    \label{fig:quartet_tree}
\end{figure}
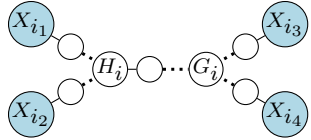

\begin{figure}[ht]
    \renewcommand{\arraystretch}{1}     
    \setlength{\tabcolsep}{5pt}  
    \begin{tabular}{ccc}
     \begin{tikzpicture}
    [
      scale=0.75,
      observed/.style={circle,inner sep=0.3mm,draw=black,fill=MyBlue}, 
      hidden/.style={circle,inner sep=0.3mm,draw=black}
    ]
    \node [observed,name=z1] at (-1.2,0.5) {$\mathsmaller X_1$};
    \node [observed,name=z2] at (-1.2,-0.5) {$\mathsmaller X_2$};
    \node [observed,name=z3] at (1.2,0.5) {$\mathsmaller X_3$};
    \node [observed,name=z4] at (1.2,-0.5) {$\mathsmaller X_4$};
    \node [hidden,name=h] at ($(-0.4,0)$) {$\mathsmaller H$};
    \node [hidden,name=g] at ($(0.4,0)$) {$\mathsmaller G$};
    \draw [-] (z1) to (h);
    \draw [-] (z2) to (h);
    \draw [-] (z3) to (g);
    \draw [-] (z4) to (g);
    \draw [-] (h) to (g);
    \end{tikzpicture}
    &
    \begin{tikzpicture}
    [
      scale=0.75,
      observed/.style={circle,inner sep=0.3mm,draw=black,fill=MyBlue},
      hidden/.style={circle,inner sep=0.3mm,draw=black}
    ]
    \node [observed,name=z1] at (-1.2,0.5) {$\mathsmaller X_1$};
    \node [observed,name=z3] at (-1.2,-0.5) {$\mathsmaller X_3$};
    \node [observed,name=z2] at (1.2,0.5) {$\mathsmaller X_2$};
    \node [observed,name=z4] at (1.2,-0.5) {$\mathsmaller X_4$};
    \node [hidden,name=h] at ($(-0.4,0)$) {$\mathsmaller H$};
    \node [hidden,name=g] at ($(0.4,0)$) {$\mathsmaller G$};
    \draw [-] (z1) to (h);
    \draw [-] (z3) to (h);
    \draw [-] (z2) to (g);
    \draw [-] (z4) to (g);
    \draw [-] (h) to (g);
    \end{tikzpicture}
    &
    \begin{tikzpicture}
    [
      scale=0.75,
      observed/.style={circle,inner sep=0.3mm,draw=black,fill=MyBlue},
      hidden/.style={circle,inner sep=0.3mm,draw=black}
    ]
    \node [observed,name=z1] at (-1.2,0.5) {${\mathsmaller X_1}$};
    \node [observed,name=z4] at (-1.2,-0.5) {$\mathsmaller X_4$};
    \node [observed,name=z2] at (1.2,0.5) {$\mathsmaller X_2$};
    \node [observed,name=z3] at (1.2,-0.5) {$\mathsmaller X_3$};
    \node [hidden,name=h] at ($(-0.4,0)$) {$\mathsmaller H$};
    \node [hidden,name=g] at ($(0.4,0)$) {$\mathsmaller G$};
    \draw [-] (z1) to (h);
    \draw [-] (z4) to (h);
    \draw [-] (z2) to (g);
    \draw [-] (z3) to (g);
    \draw [-] (h) to (g);
    \end{tikzpicture}\\
    $\{\{1,2\},\{3,4\}\}$
    & $\{\{1,3\},\{2,4\}\}$
    &  $\{\{1,4\},\{2,3\}\}$
    \end{tabular}
    \centering
    \caption{Three fixed ways to connect $X_1$, $X_2$, $X_3$, $X_4$, with two latent variables $H$ and $G$.}
    \label{fig:topologies}
\end{figure}
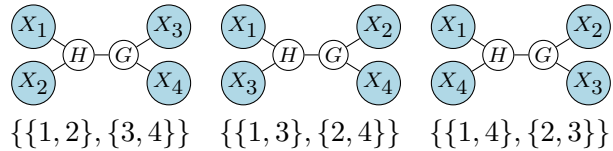

{\bf Quartet.} A quadruple of observed variables from a latent tree $\Tcal$ is called a quartet (Figure~\ref{fig:quartet_tree}).
Under assumption {\bf (A2)}, there are $3$ ways to connect a quartet, $X_1, X_2, X_3$, $X_4$, using $2$ latent variables $H$ and $G$ (Figure~\ref{fig:topologies}).
However, only one of the 3 quartet relations is consistent with $\,\,\Tcal$. The mapping between quartets and the tree topology 
$\Tcal$  is captured in the following theorem~\citep{Buneman71}:
\begin{theorem} 
\label{th:quartets_and_tree}
  The set of all quartet relations $\Qcal_{\Tcal}$ is unique to a latent tree $\Tcal$, and furthermore, $\Tcal$ can be recovered from $\Qcal_{\Tcal}$ in polynomial time.
\end{theorem}

{\bf Quartet-based tree reconstruction.} Motivated by Theorem~\ref{th:quartets_and_tree}, a family of latent tree recovery algorithms has been designed based on resolving quartet relations. These algorithms first determine one of the $3$ ways how $4$ variables are connected, and then join together all quartet relations to form a consistent latent tree. For a model with $d$ observed variables, there are $O(d^4)$ quartet relations in total (taking all possible combinations of $4$ variables). However, we do not necessarily need to resolve all these quartet relations in order to reconstruct the latent tree. A small set of size $O(d\log d)$ will suffice for the tree recovery, which makes quartet based methods efficient even for problems with large $d$~\citep{PeaTar86,Pearl88}. In this paper, we design a new quartet based method. Our main contribution compared to previous approaches is that our method is \emph{agnostic} to the number of hidden states, $k$, which is usually unknown in practice.

\section{Resolving Quartet Relations without Knowing the Number of Hidden States}

\begin{figure}[ht]
\centering
    \psfrag{1}{\hspace*{-2mm}\begin{footnotesize}$P_{1|H}$\end{footnotesize}}
    \psfrag{2}{\hspace*{-3mm}\begin{footnotesize}$P_{2|H}$\end{footnotesize}}
    \psfrag{3}{\hspace*{-2mm}\begin{footnotesize}${\cal I}_H$\end{footnotesize}}
    \psfrag{5}{\hspace*{-2mm}\begin{footnotesize}$P_{HG}$\end{footnotesize}}
    \psfrag{6}{\begin{footnotesize}${\cal I}_G$\end{footnotesize}}
    \psfrag{7}{\hspace*{-3mm}\begin{footnotesize}$P_{4|G}$\end{footnotesize}}
    \psfrag{8}{\hspace*{-3mm}\begin{footnotesize}$P_{3|G}$\end{footnotesize}}
    \includegraphics[width=.3\textwidth]{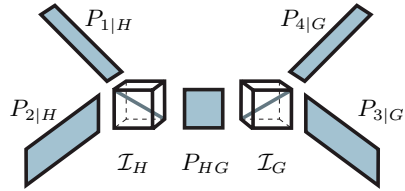} 
  \caption{Schematic diagram of the tensor $\Pcal(X_1,X_2,X_3,X_4)$.}
  \label{fig:tensor}
\end{figure}

In this section, we develop a test for resolving the latent relation of a quartet when the number of hidden states is unknown. Our approach makes use of information from the joint probability table of a quartet, which is a $4$-way table or $4$th order tensor. Suppose that the quartet relation of $4$ variables, $X_1, X_2, X_3$ and $X_4$, is $\{\{1,2\},\{3,4\}\},$ then the entries in this tensor are specified by
\begin{align}
  \Pcal&(x_1, x_2, x_3, x_4) =
  \sum\nolimits_{h, g} P(x_1 | h) P(x_2 | h) P(h,g) P(x_3|g) P(x_4 | g).
\label{def:P}
\end{align}
This factorization suggests that there exist some low rank structures in the $4$th order tensor. 
To study the rank properties of $\Pcal(X_1,X_2,X_3,X_4)$, we first relate it to the conditional probability tables, $P(X_1|H)$, $P(X_2|H)$, $P(X_3|G)$, $P(X_4|G)$, and the joint probability table, $P(H,G)$ (we abbreviate them as $ P_{1|H}$, $P_{2|H}$, $P_{3|G}$, $P_{4|G}$ and $P_{HG}$, respectively). Using tensor algebra, we have
$${\cal P}(X_1,X_2,X_3,X_4)= \langle\Tcal_1 , \Tcal_2\rangle_3,$$
 $$\begin{array}{ll}
   \mbox{with} & \Tcal_1 = \Ical_H \times_1 P_{1|H} \times_2 P_{2|H},\\[1mm]
  & \Tcal_2 = \Ical_G \times_1 P_{3|G} \times_2 P_{4|G} \times_3 P_{HG},
  \end{array}$$
where ${\cal I}_H$ and ${\cal I}_G$ are $3$rd order diagonal tensors of size $k\times k \times k$ with diagonal elements equal to $1$. The multiplication $\times_i$ denotes a tensor-matrix multiplication with respect to the $i$-th dimension of the tensor and the rows of the matrix, and 
$\langle\cdot,\cdot\rangle_3$ denotes tensor-tensor multiplication along the third dimension of both tensors\footnote{For formal definitions of tensor notations see appendix, \S\ref{sect:properties}.}. This formula can be schematically understood as Figure~\ref{fig:tensor}. 
We will start by characterizing the rank properties of ${\cal P}$ and then exploit them to design a quartet test. Although the proposed approach involves unfolding the tensor and subsequent computation at the matrix level, modeling the problem using tensors provides higher level conceptual understanding of the structure of ${\Pcal}.$ The novelty of our use of low rank tensors is for latent structure discovery. 

\subsection{Unfolding the $4$th Order Tensor}
\begin{figure}[ht]
\centering
  \begin{tabular}{cc}
  \hspace*{-20mm}
    \psfrag{0}{\hspace*{-2mm}\begin{footnotesize}$~$\end{footnotesize}}
    \psfrag{1}{\hspace*{-2mm}\begin{footnotesize}$P_{2|H}$\end{footnotesize}}
    \psfrag{2}{\hspace*{-2mm}\begin{footnotesize}$P_{1|H}$\end{footnotesize}}
    \psfrag{4}{\hspace*{-2.6mm}\begin{footnotesize}$P_{HG}$\end{footnotesize}}
    \psfrag{5}{\hspace*{-2mm}\begin{footnotesize}$P_{4|G}$\end{footnotesize}}
    \psfrag{6}{\hspace*{-2mm}\begin{footnotesize}$P_{3|G}$\end{footnotesize}}
    \psfrag{T}{\begin{tiny}$\top$\end{tiny}}
    \includegraphics[width=.40\textwidth]{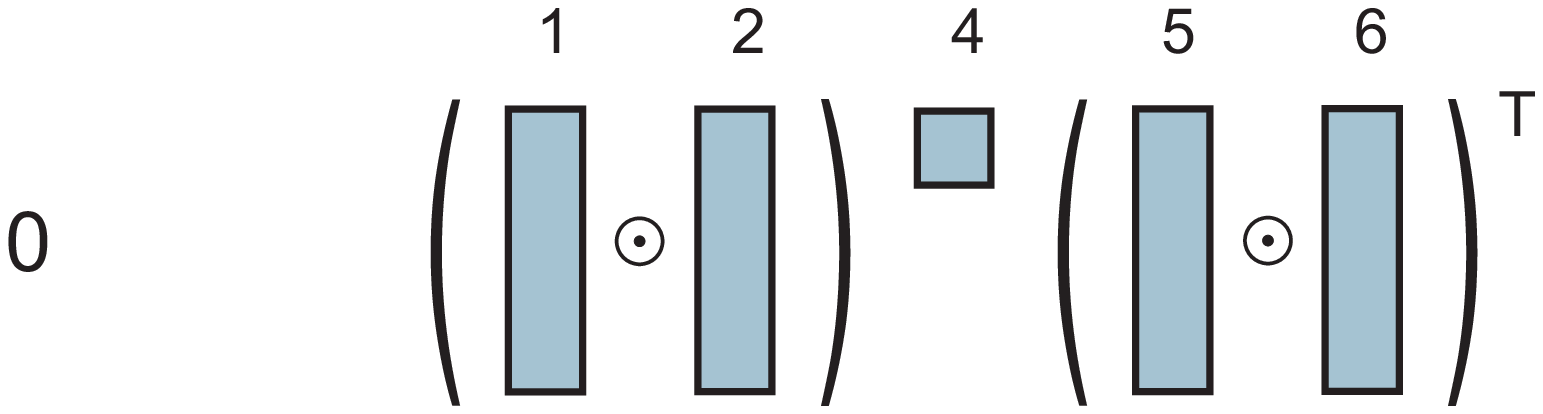} 
    &
    \hspace*{-5mm}
    \psfrag{0}{\hspace*{-2mm}\begin{footnotesize}$~$\end{footnotesize}}
    \psfrag{1}{\hspace*{-3mm}\begin{footnotesize}$P_{3|G}$\end{footnotesize}}
    \psfrag{2}{\hspace*{-3mm}\begin{footnotesize}$P_{1|H}$\end{footnotesize}}
    \psfrag{3}{\hspace*{-7mm}\begin{footnotesize}$\mathsmaller{\diag}(P_{HG}(:))$\end{footnotesize}}
    \psfrag{5}{\hspace*{-1.5mm}\begin{footnotesize}$P_{4|G}$\end{footnotesize}}
    \psfrag{6}{\hspace*{-1mm}\begin{footnotesize}$P_{2|H}$\end{footnotesize}}
    \psfrag{T}{\begin{tiny}$\top$\end{tiny}}
    \includegraphics[width=.40\textwidth]{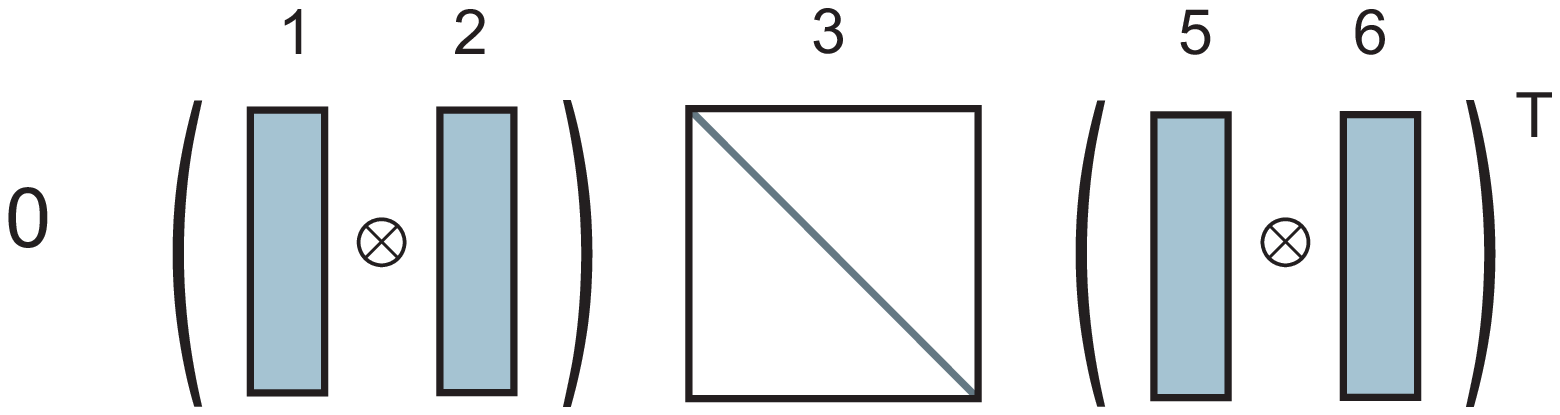} \\
    \hspace*{6mm}(a)\hspace*{2mm} \begin{footnotesize}$A$\end{footnotesize} 
    & \hspace*{4mm}(b)\hspace*{2mm} \begin{footnotesize}$B$\end{footnotesize}  
  \end{tabular}
  \caption{Schematic diagrams of the two unfoldings $A$ and $B$.}
  \label{fig:ABcompact}
\end{figure}

Now we consider 3 different reshapings  $A,\,B$ and $C$ of the tensor into matrices (``unfoldings''). These unfoldings contain exactly the same entires as $\Pcal$ but in different order.  $A$ corresponds to the grouping $\{\{1,2\},\{3,4\}\}$ of the variables, \ie, the rows of $A$ correspond to dimensions $1$ and $2$ of $\Pcal$, and its columns to dimensions $3$ and $4$. $B$ corresponds to the grouping $\{\{1,3\},\{2,4\}\}$ and $C$ - to the grouping $\{\{1,4\},\{2,3\}\}$. Using {\sc Matlab}'s notation (see appendix, \S\ref{sect:properties} for further explanation), \newpage\vspace*{-1.2cm}
\begin{align}
  \label{def:A} A & = \mbox{reshape}({\Pcal},n^2,n^2);\\
  \label{def:B} B & = \mbox{reshape}(\mbox{permute}({\Pcal},[1, 3, 2, 4]),n^2,n^2);\\
  \label{def:C} C & = \mbox{reshape}(\mbox{permute}({\Pcal},[1, 4, 2, 3]),n^2,n^2).
\end{align}
Next we present useful characterizations of
$A,\,B$ and $C$, which will be essential for understanding their connection with the latent structure of a quartet. The {\it Kronecker product} of two matrices $M$ and $M'$ is denoted as $M\otimes M'$, and if they have the same number of columns, their {\it Khatri-Rao product}  (column-wise Kronecker product), is denoted as $M\odot M'$. Then (see appendix \S\ref{sect:fromP_toABC} for proof),
\begin{lemma} 
  Assume that $\{\{1,2\},\{3,4\}\}$ is the correct latent structure. The matrices $A$, $B$ and $C$ can be factorized respectively as 
  (see Figure~\ref{fig:ABcompact}(a) and Figure~\ref{fig:ABcompact}(b) for schematic diagrams)
  \begin{align}
    \hspace*{-1mm}A &= \big(P_{2|H} \odot P_{1|H}\big)\,\,\, P_{HG}   \,\,\, \big(P_{4|G} \odot P_{3|G}\big)^\top, \label{eq:Acompact} \\
    \hspace*{-1mm}B &= \big(P_{3|G} \otimes P_{1|H}\big)\,\diag(P_{HG}(:))\,\big(P_{4|G} \otimes P_{2|H}\big)^\top, \label{eq:Bcompact} \\
    \hspace*{-1mm}C &= \big(P_{4|G} \otimes P_{1|H}\big)\,\diag(P_{HG}(:))\,\big(P_{3|G} \otimes P_{2|H}\big)^\top. \label{eq:Ccompact}
  \end{align}
  \label{le:unfolding}
\end{lemma}\vspace*{-5mm}

The factorization of $A$ is very different from those of $B$ and $C$. First, in $A$, $P_{2|H}\odot P_{1|H}$ is a matrix of size $n^2\times k$, and the columns of $P_{2|H}$ interact only with their corresponding columns in $P_{1|H}$. However, in $B$, $P_{3|G}\otimes P_{1|H}$ is a matrix of size $n^2\times k^2$, and every column of $P_{1|H}$ interacts with every column of $P_{3|G}$ respectively (similarly for $C$). Second, in $A$, the middle factor $P_{HG}$ has size $k\times k$, whereas in $B$, the entires of $P_{HG}$ appear as the diagonal of a matrix of size $k^2\times k^2$ (similarly for $C$). These differences result in different rank properties of $A,\,B$ and $C$ which we will exploit to discover the latent structure of a quartet.

\subsection{Rank Properties of the Unfoldings}
\label{sect:rank_properties}
Under assumption {\bf (A1)} that all CPTs have full column rank, the factorization of $A$, $B$ and $C$ in~\eq{eq:Acompact},~\eq{eq:Bcompact} and~\eq{eq:Ccompact} respectively suggest that (see appendix \S\ref{sect:fromP_toABC} for more details)
\begin{align}
  \text{rank}(A) = \text{rank}(P_{HG}) = k
  ~\leq~ \text{rank}(B) = \text{rank}(C) = \text{nnz}(P_{HG}), \label{eq:rankA_}
\end{align}
where $\text{nnz}(\cdot)$ denotes the number of nonzero elements.
We note that the equality is attained if and only if the relationship between the hidden variables $G$ and $H$
is deterministic, \ie, there is a single nonzero element in each row and in each column 
of $P_{HG}$. In this case, the grouping of variables in a quartet can be arbitrary, and we will not consider this case in the paper. More specifically, we have
\begin{theorem}
  Assume $P_{HG}$ has a few zero entries, then
  $k \ll k^2 \approx \,\mbox{\textnormal{nnz}}(P_{HG})$
  and thus
  \begin{align}
    {\boxed{
    \mbox{\textnormal{rank}}(A) \ll \mbox{\textnormal{rank}}(B) = \mbox{\textnormal{rank}}(C).}
    \label{eq:rank_condition}}
  \end{align}
  \label{th:rank}
\end{theorem}\newpage
The above theorem reveals a useful difference between the correct grouping of variables 
and the two incorrect ones. Furthermore, this condition can be easily verified:
Given $\Pcal$ we can check the rank of its matrix representations
$A,\,B$ and $C$ and thus discover the latent structure of the quartet.

\subsection{Nuclear Norm Relaxation for the Rank Condition}
In practice, due to sampling noise all unfolding matrices $A,\,B$ and $C$ would
be nearly full rank, so the rank condition cannot be applied directly. 
To deal with this, we design a test based on relaxation of the rank condition using nuclear norm 
\begin{align}
  \|M\|_\ast = \sum\nolimits_{i=1}^{n} \sigma_i(M), 
\end{align}
which is the sum of all singular values of an $(n\times n)$ matrix $M$. Instead of comparing the ranks of $A,\,B$ and $C$, we look for the one with the smallest nuclear norm and declare the latent structure corresponding to it. This simple quartet algorithm is summarized in Algorithm~\ref{alg:main}.
\begin{algorithm}[htb]
  \caption{$i^\ast=$ Quartet($X_1$, $X_2$, $X_3$, $X_4$)}
  \begin{algorithmic}[1] 
  \STATE Estimate $\widehat{\Pcal}(X_1,X_2,X_3,X_4)$ from a set of $m$~\iid~samples $\{(x_1^l, x_2^l, x_3^l, x_4^l)\}_{l=1}^{m}$.\;\\
  \STATE Unfold $\widehat{\Pcal}$ in three different ways into matrices $\widehat{A}$, $\widehat{B}$ and $\widehat{C}$, and compute their nuclear norms\\
    \hspace*{4mm}$a_1 = \|\widehat{A}\|_\ast,~a_2 = \|\widehat{B}\|_\ast$ and $a_3 = \|\widehat{C}\|_\ast$.\\
  \STATE Return $i^\ast = \argmin\nolimits_{i\in\{1,2,3\}} a_i$.
  \end{algorithmic}
\label{alg:main}
\end{algorithm}
Note that Algorithm~\ref{alg:main} works even if the number of hidden states, $k$, is a priori unknown.  
This is an important advantage over the idea of learning the structure 
based on additive distance~\citep{Lake1994}, where $k$ is assumed to be the same as the number of states, $n$, of the observed variables, or over a recent approach based on quartet test~\citep{AnaChaHsuKakSonZha2011}, where $k$ needs to be specified in advance.

In our current context, nuclear norm has a few useful properties. 
First, it is the tightest convex lower bound of the rank of a matrix~\citep{FazHinBoy01}.
This is why\footnote{Note that $A$, $B$ and $C$ consist of the same elements so their Frobenius norms are the same, \ie, the $3$ matrices are readily equally ``normalized''.} it is meaningful to compare nuclear norms instead of ranks.
Second, it is easy to compute: a standard singular value decomposition will
do the job. Third, it is robust to estimate. The nuclear norm of a probability matrix $\widehat{A}$~based on samples is nicely concentrated around its population quantity~\citep{RosBelVit2010}. Given a confidence level $1-2e^{-\tau}$, an estimate based on $m$ samples satisfies
\begin{align}
 |& \|A\|_\ast - \|\widehat{A}\|_\ast | = 
 \abr{\sum\nolimits_i \sigma_i(A) - \sum\nolimits_i \sigma_i(\widehat{A})} \leq 2\sqrt{2\tau}/\sqrt{m}.
 \label{eq:samplebound}
\end{align}
Fourth, the nuclear norm can be viewed as a measure of dependence between two pairs of variables. For instance, if $A$ corresponds to grouping $\{\{1,2\},\{3,4\}\}$, $\|A\|_\ast$ measures the dependence between the compound variables $\{X_1,X_2\}$ and $\{X_3,X_4\}$. In the community of kernel methods, $A$ is treated as a cross-covariance operator between $\{X_1,X_2\}$ and $\{X_3,X_4\}$, and its spectrum 
has been used to design various dependence measures, such as Hilbert-Schmidt Independence Criterion, which is the sum of squares of all singular values~\citep{GreBouSmoSch05}, and kernel constrained covariance, which only takes the largest singular value~\citep{GreHerSmoBouetal05}. Intuitively, our quartet test says that: if we group the variables correctly, then cross group dependence should be low, since the groups are separated by two latent variables; however if we group the variables incorrectly, then cross group dependence should be high, since similar variables exist in the two groups. 

\section{Recovery Conditions and Finite Sample Guarantee for Quartets} 
\label{sec:nuclearnormconditions}
Since nuclear norm is just a convex lower bound of the rank, there might be situations where the nuclear norm does not satisfy the same relation as the rank. That is, it might happen that 
$\mbox{rank}(A) \leq \mbox{rank}(B)$ but $\|A\|_\ast \geq \|B\|_\ast$. In this section, we present sufficient conditions under which nuclear norm returns successful quartet test. 

{\bf When latent variables $H$ and $G$ are independent}, rank$(P_{HG})=1$,
since $P_{HG} = P_H P_G^\top$ ($P(h,g)=P(h)P(g)$). Let $\{\{1,2\},\{3,4\}\}$ be the correct quartet relation. We can obtain simpler characterizations of the 3 unfoldings of $\Pcal(X_1,X_2,X_3,X_4)$, denoted as $A_{\perp}$, $B_{\perp}$ and $C_{\perp}$ respectively. Using Lemma~\ref{le:unfolding} and the independence of $H$ and $G$, we have (see appendix, (\ref{eq:B_perp_nn})--(\ref{eq:A_perp_nn}))
\begin{equation}
  \begin{array}{llcl}
    \hspace*{-2mm}A_\perp\hspace*{-2mm}
    & = (P_{2|H} \odot P_{1|H})\,\,\, P_H P_G^\top \,\,\, (P_{4|G} \odot P_{3|G})^\top\\[1mm]
    & = P_{12}(:)~P_{34}(:)^\top, \\[2mm]
    \hspace*{-2mm}B_\perp\hspace*{-2mm}
    & = (P_{3|G} \otimes P_{1|H}) ({\mathsmaller\diag}(P_G) \otimes {\mathsmaller\diag}(P_H)) (P_{4|G}\otimes P_{2|H})^\top\\[1mm]
    & = P_{34} \otimes P_{12},
  \end{array}
  \label{eq:B_perp}
\end{equation}
and $\mbox{rank}(A_{\perp})=1 \ll \rank(B_{\perp})$ which is consistent with Theorem~\ref{th:rank}. Furthermore, since $A_{\perp}$ has only one nonzero singular value, we have $\|A_{\perp}\|_\ast = \|A_{\perp}\|_F = \|B_{\perp}\|_F \leq \|B_{\perp}\|_\ast$ (using $\|M\|_F \leq \|M\|_\ast$ for any matrix $M$). Similarly, $C_\perp=P_{43} \otimes P_{12}$ and $\|A_{\perp}\|_\ast  \leq \|C_{\perp}\|_\ast$. Then we know for sure that the nuclear norm quartet test will return the correct topology.

{\bf When latent variables $H$ and $G$ are not independent}, we treat it as perturbation $\Delta$ away from the independent case,~\ie,~$\widetilde{P}_{HG} = P_H P_G^\top + \Delta$. The size of $\Delta$ quantifies the strength of dependence between $H$ and $G$. Obviously, when $\Delta$ is small,~\eg,~$\Delta=\zero$, we are back to the independence case and it is easy to discover the correct quartet relation; when it is large,~\eg,~$\Delta = I - P_H P_G^\top$, $H$ and $G$ are deterministically related and the different groupings are indistinguishable. The question is how large can $\Delta$ be while still allowing the nuclear norm quartet test to find the correct latent relation.  

First, we require {\bf (A3)} $\Delta \one = \zero$, and $\Delta^\top \one = \zero$, where $\one$ and $\zero$ are vectors of all ones and all zeros. Such perturbation $\Delta$ keeps the marginal distributions $P_H$ and $P_G$ as in the independent case, since $\widetilde{P}_H=\widetilde{P}_{HG} \one = P_H P_G^\top \one + \Delta \one = P_H$. Assuming $\{\{1,2\},\{3,4\}\}$ is the correct quartet relation, $\Delta$ also keeps the pairwise marginal distribution $P_{12}$ as in the independent case, since $P_{12} = P_{1|H} \diag(P_H) P_{2|H}^\top$ and the marginal $P_H$ is the same before and after the perturbation. Similar reasoning also applies to $P_{34}=P_{3|G} \diag(P_G) P_{4|G}^\top$.
 
We define \emph{excessive dependence} of the correct and incorrect groupings as
$$\theta := \min \{\|B_\perp\|_\ast - \|A_\perp\|_\ast,~\|C_\perp\|_\ast - \|A_\perp\|_\ast\}.$$
It quantifies the changes in dependence when we switch from incorrect groupings to the correct one
(in the case when $H$ and $G$ are independent). Note that $\theta$ is measured only from pairwise marginals (\ref{eq:B_perp}), $P_{12}$ and $P_{34}$. Using matrix perturbation analysis we can show that (see appendix $\S$\ref{sect:perturbation} for proof)
\begin{lemma}
  \label{le:deltacondition}
  If $\nbr{\Delta}_F \leq \frac{\theta}{{k^2} +k}$, then Algorithm~\ref{alg:main} returns the correct quartet relation.
\end{lemma}
Thus, if the excessive dependence $\theta$ is large compared to the number of hidden states, the size of the allowable perturbation can be correspondingly larger. In other words, if the dependence between variables within the same group is strong enough compared to the dependence across groups, we allow for larger $\Delta$ and stronger dependence between hidden variables $H$ and $G$ (which is closer to the indistinguishable case).
Then under the recovery condition in Lemma~\ref{le:deltacondition}, and given $m$~\iid~observations, we can obtain the following guarantee for the quartet test (see appendix, $\S$\ref{app:stat:quartet} for proof). Let $ \alpha = \min \cbr{\|B\|_\ast - \|A\|_\ast, \|C\|_\ast - \|A\|_\ast}$.
\begin{lemma}
  \label{le:quartetsuccess}
  With probability $1-8 e^{-\frac{1}{32}m\alpha^2}$, Algorithm~\ref{alg:main} returns the correct quartet relation.
\end{lemma}

\section{Building Latent Tree from Quartets}

{\bf Algorithm.} We can  use the resolved quartet relations (Algorithm~\ref{alg:main}) to discover the structure of the entire tree
via an incremental divide-and-conquer algorithm~\citep{PeaTar86,Pearl88}, summarized in Algorithm~\ref{alg:buildtree}
(further details in appendix \S\ref{app:build_tree}).
Joining variable $X_{i+1}$ to the current tree of $i$ leaves can be done with 
$O(\log i)$ tests. This amounts to performing 
$O(d\log d)$ quartet tests for building an entire tree of $d$ leaves, which is efficient even if $d$ is large.
Moreover, as shown in~\citep{PeaTar86}, this algorithm is consistent.
\begin{algorithm}[ht]
\caption{${\Tcal}$ = BuildTree$(X_1,\ldots, X_d)$}
 \begin{algorithmic}[1] 
\STATE Connect any $4$ variables $X_1$, $X_2$, $X_3$, $X_4$ with $2$ latent variables in a tree $\Tcal$ using Algorithm~\ref{alg:main}.
\FOR[insert $\mathsmaller{(i+1)}$-th leaf $X_{i+1}$]{$i=4,5,\ldots,d-1$} 
\STATE Choose root $R$ that splits $\Tcal$ into sub-trees $\Tcal_1,\Tcal_2,\Tcal_3$ of roughly equal size.
\STATE Choose any triplet $(X_{i_1},X_{i_2},X_{i_3})$ of leaves from different sub-trees.
\STATE Test which sub-tree should  $X_{i+1}$ be joined to:\\
   $i^\ast \leftarrow$ Quartet($X_{i+1},X_{i_1},X_{i_2},X_{i_3}$).
\STATE Repeat recursively from step 3 with ${\Tcal} := {\Tcal}_{i^\ast}$.\\
  This will eventually reduce to a tree with a single leaf. Join $X_{i+1}$ to it via hidden variable. 
\ENDFOR
\end{algorithmic}
\label{alg:buildtree}
\end{algorithm}

{\bf Tree recovery conditions and guarantees.} How will the quartet recovery conditions translate to recovery conditions for the entire tree,
where each ``edge'' of a quartet is a path in the tree? What are the finite sample guarantees for the divide-and-conquer algorithm? 

When a quartet is taken from a latent tree, each edge of the quartet corresponds to a path in the tree involving a chain of variables (Figure~\ref{fig:topologies}). We need to bound the perturbation to each single edge of the tree such that joint path perturbations
satisfy edge perturbation conditions from Lemma~\ref{le:deltacondition}. For a quartet $q=\{\{i_1,i_2\},\{i_3,i_4\}\}$ corresponding to a single edge between $H$ and $G$, denote the excessive dependence by $\theta_q$. 
By adding perturbation $\Delta_q$ of size smaller than $\frac{\theta_q}{k^2+k}$ to $P_H P_G^\top$ we can still correctly recover $q$. Let $\theta_{\min}:=\min_{\text{quartet}~ q}\theta_q$. If we require $\|\Delta_q\|_F \leq \frac{\theta_{\min}} {k^2+k}$, all such quartet relations will be recovered successfully. If we further restrict the size of the perturbation by 
the smallest value in a marginal probability distribution of a hidden variable, $\gamma_{\min}:=\min_{\text{hidden node}~H} \min_{i=1\ldots k} P_H(i)$, we can guarantee that all quartet relations corresponding to a path between $H$ and $G$ can also be successfully recovered by the nuclear norm test (see appendix \S\ref{app:recovery_tree}). Therefore, we assume that {\bf (A4)} $\nbr{\Delta_q}_F \leq \min\{\frac{\theta_{\min}}{{k^2} +k},\gamma_{\min}\}$ for all quartets $q$ in a tree.
\begin{theorem}
  \label{th:treecondition}
  Algorithm~\ref{alg:buildtree} returns the correct tree topology under assumptions {\bf (A1)--(A4)}.
\end{theorem}
The recovery conditions guarantee that all quartet relations can be resolved  correctly and simultaneously. Then a consistent algorithm using a subset of the quartet relations should return the correct tree structure. Given $m$~\iid~samples, we have the following statistical guarantee for the tree building algorithm (see appendix, $\S$\ref{app:stat:tree} for proof). Let $\alpha_{\min}:=\min_{\text{quartet}~q}\alpha_q$.
\begin{theorem}
 With probability $1-8\cdot c\cdot d\log d \cdot e^{-\frac{1}{32}m\alpha_{\min}^2}$, 
 Algorithm~\ref{alg:buildtree} recovers the correct tree topology for a constant $c$ under assumptions {\bf (A1)--(A4)} . 
\end{theorem}
We note that there are better quartet based algorithms for building latent trees with stronger statistical guarantees,~\eg~\citep{ErdSzeSteWar99b}. We can adapt our nuclear norm based quartet test to those algorithm as well. However, this is not the main focus of the paper. We choose the divide-and-conquer algorithm due to its simplicity, ease of analysis and it illustrates well how our quartet recovery guarantee can be translated into a tree building guarantee. 

\section{Experiments}
\label{sec:experiments}

\newcommand{\nj}{{NJ}}
\newcommand{\spectralk}{{Spectral@$k$}}

We compared our algorithm with representative algorithms: the neighbor-joining algorithm (\nj) \citep{SaiNei87}, a quartet based algorithm of~\citet{AnaChaHsuKakSonZha2011} ({\spectralk}), the Chow-Liu neighbor Joining algorithm (CLNJ)~\citep{Choi11}, and an algorithm of~\citet{HarWil10}~ (HW).

{\nj}~proceeds by recursively joining two variables that are closest according to an additive distance defined as $
  d_{ij} = \smallfrac{1}{2} \log \det \diag P_i - \log |\det P_{ij}| + \smallfrac{1}{2} \log \det \diag P_j, 
$
where ``det'' denotes determinant, ``diag'' is a diagonalization operator, $P_{ij}$ denotes the joint probability table $P(X_i,X_j)$, and $P_i$ and $P_j$ the probability vector $P(X_i)$ and $P(X_j)$ respectively~\citep{Lake1994}. When $P_{ij}$ has rank $k < n$, $\log |\det P_{ij}|$ is not defined,~\nj~can perform poorly. {\spectralk}~uses singular values of $P_{ij}$ to design a quartet test~\citep{AnaChaHsuKakSonZha2011}. For instance, if the true quartet configuration is $\{\{1,2\},\{3,4\}\}$ as in Figure~\ref{fig:topologies}, then the quartet needs to satisfy
$\prod\nolimits_{s=1}^k \sigma_s(P_{1 2}) \sigma_s(P_{34})>\max\{\prod\nolimits_{s=1}^k \sigma_s(P_{1 3}) \sigma_s(P_{2 4}),~
   \prod\nolimits_{s=1}^k \sigma_s(P_{1 4}) \sigma_s(P_{2 3})\} 
$. Based on this relation, a confidence interval based quartet test is designed and used as a subroutine for a tree reconstruction algorithm. {\spectralk} can handle cases with $k < n$, but still require $k$ as an input. We will show in later experiments that its performance is sensitive to the choice of $k$.~CLNJ first applies Chow-Liu algorithm~\citep{ChowLiu68} to obtain a fully observed tree and then proceeds by adding latent variables using neighbor joining algorithm. The HW algorithm is a greedy algorithm to learn binary trees by iteratively joining two nodes with a high mutual information. The number of hidden states is automatically determined in the HW algorithm and can be different for different latent variables. 

\subsection{Resolving Quartet Relations}
\begin{figure*}[t!]
  \centering
  \subfigure[$k=\{2,4\}, \mu=0.5$]{\label{fig:quartet:a}\includegraphics[width=.25\textwidth]{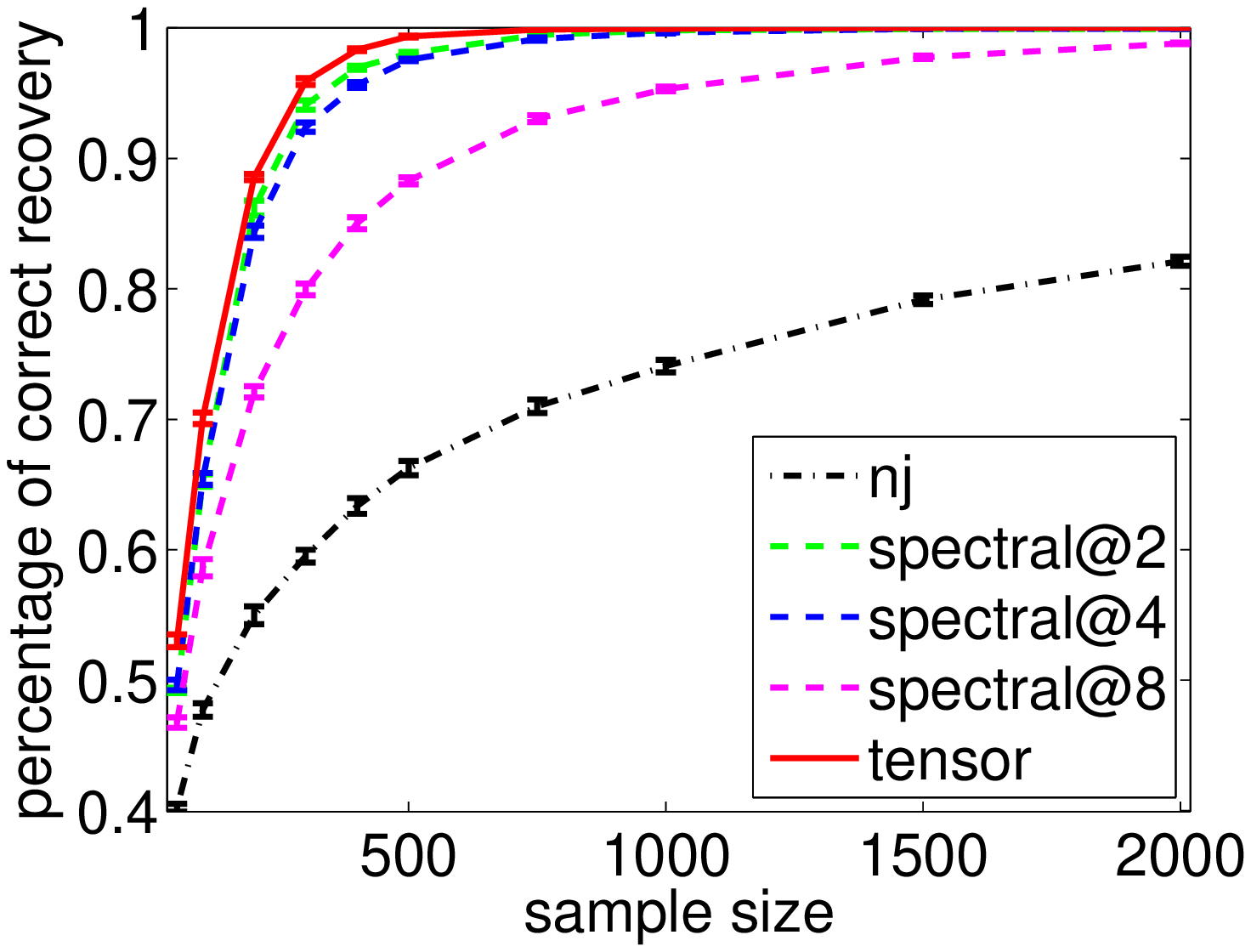}}
  \subfigure[$k=\{2,8\}, \mu=0.5$]{\label{fig:quartet:b}\includegraphics[width=.25\textwidth]{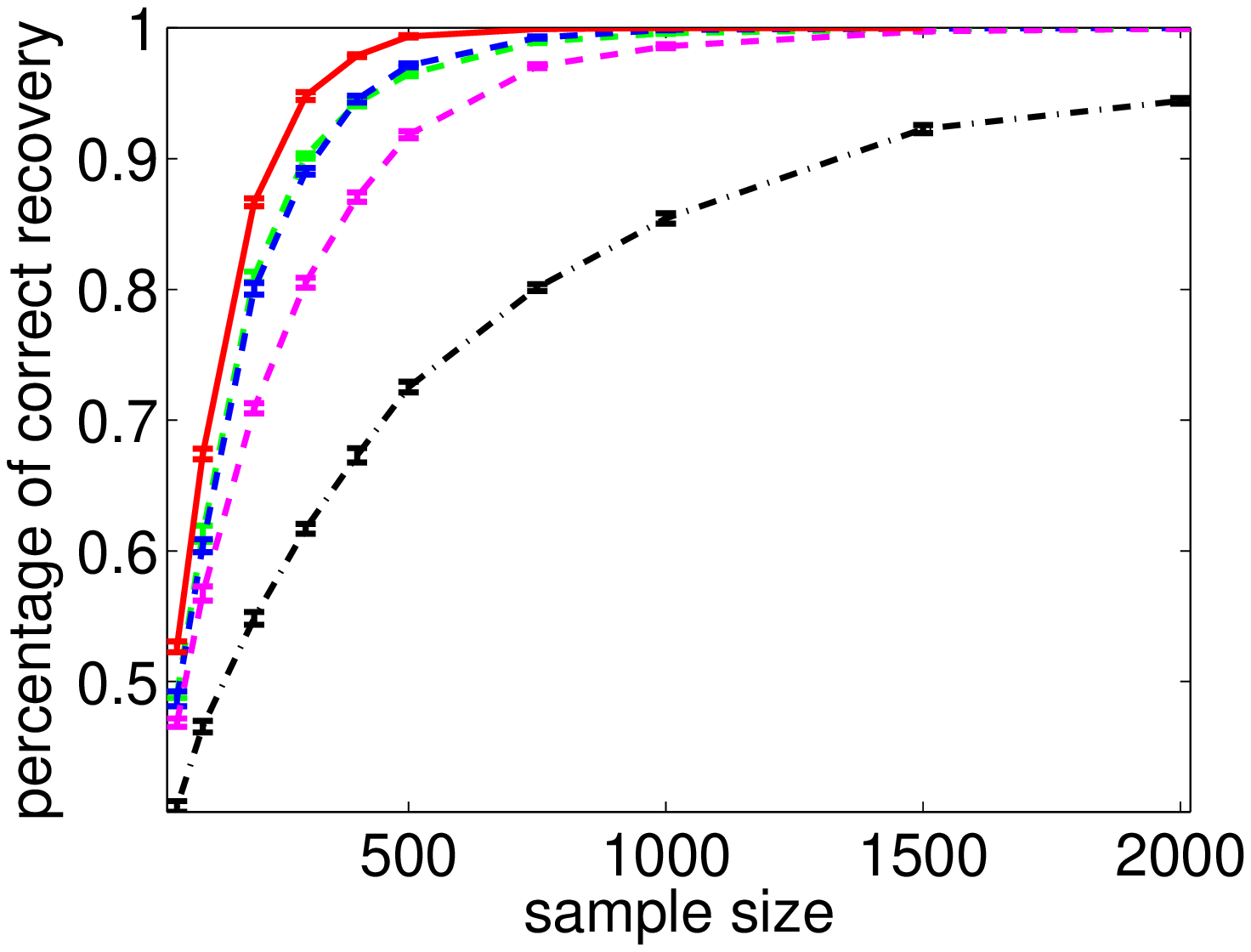}}
  \subfigure[$k =\{4,4\}, \mu=0.5$]{\label{fig:quartet:c}\includegraphics[width=.25\textwidth]{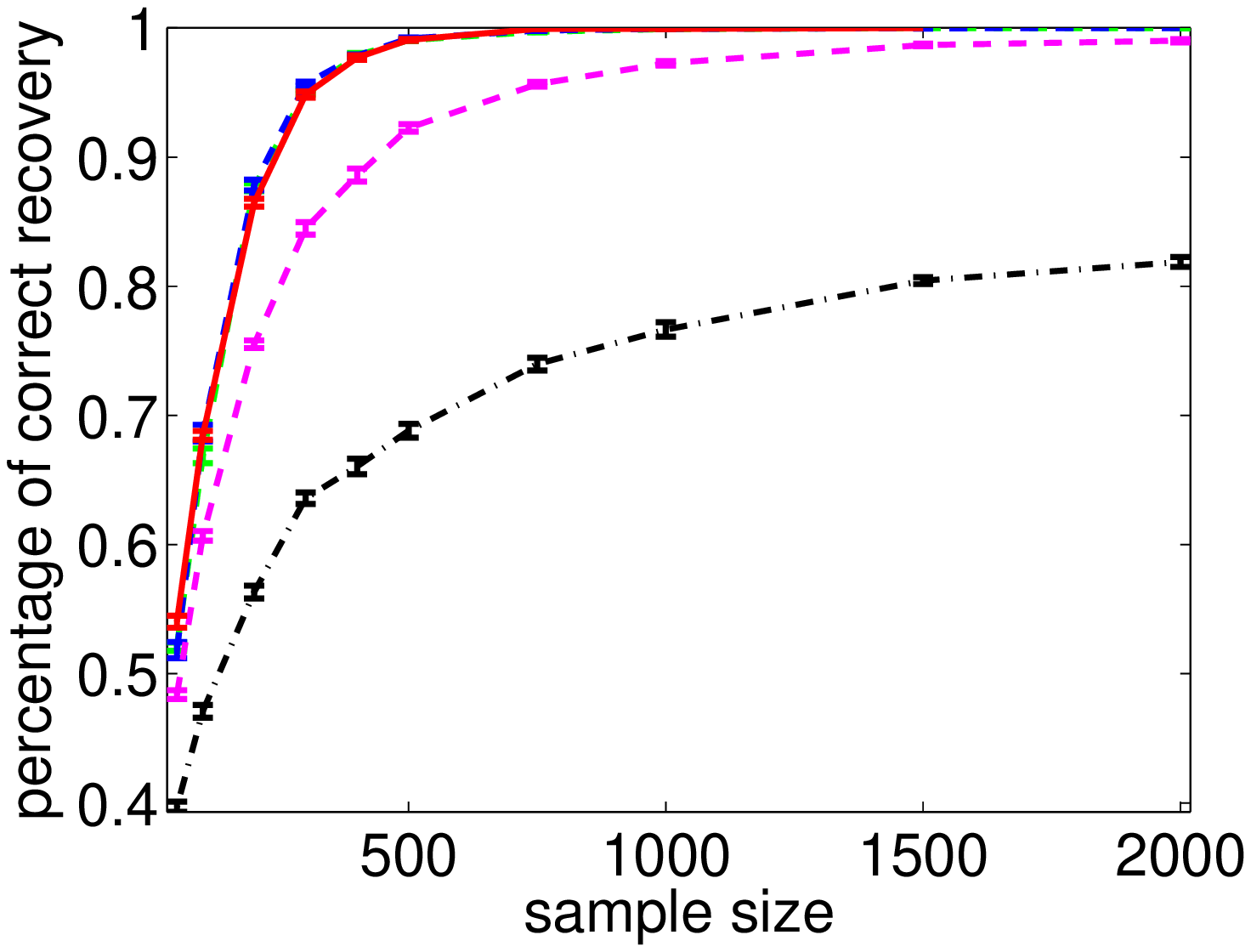}}
  \\ 
  \subfigure[$k=\{2,4\}, \mu=1$]{\label{fig:quartet:d}\includegraphics[width=.25\textwidth]{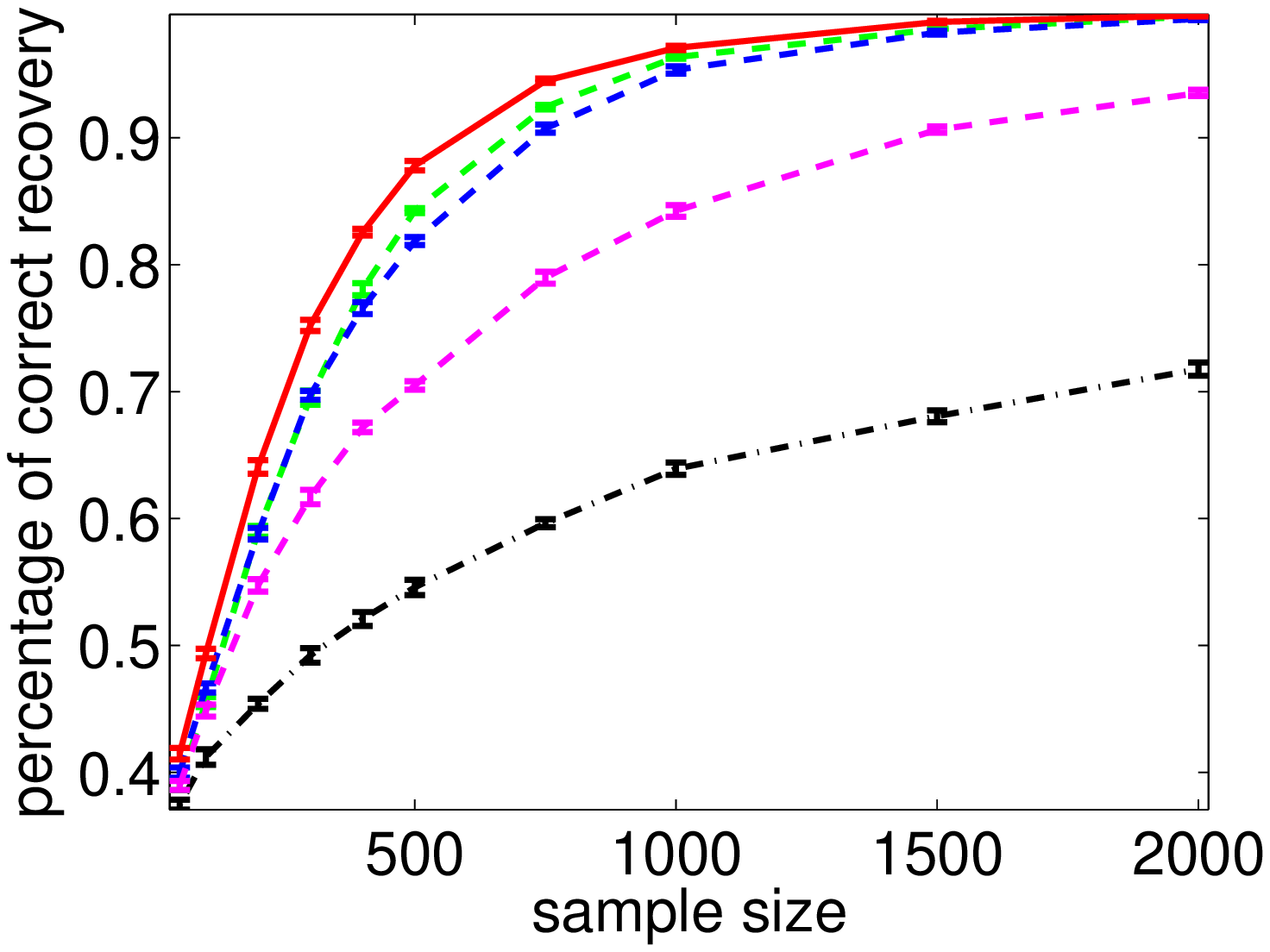}}
  \subfigure[$k=\{2,8\}, \mu=1$]{\label{fig:quartet:e}\includegraphics[width=.25\textwidth]{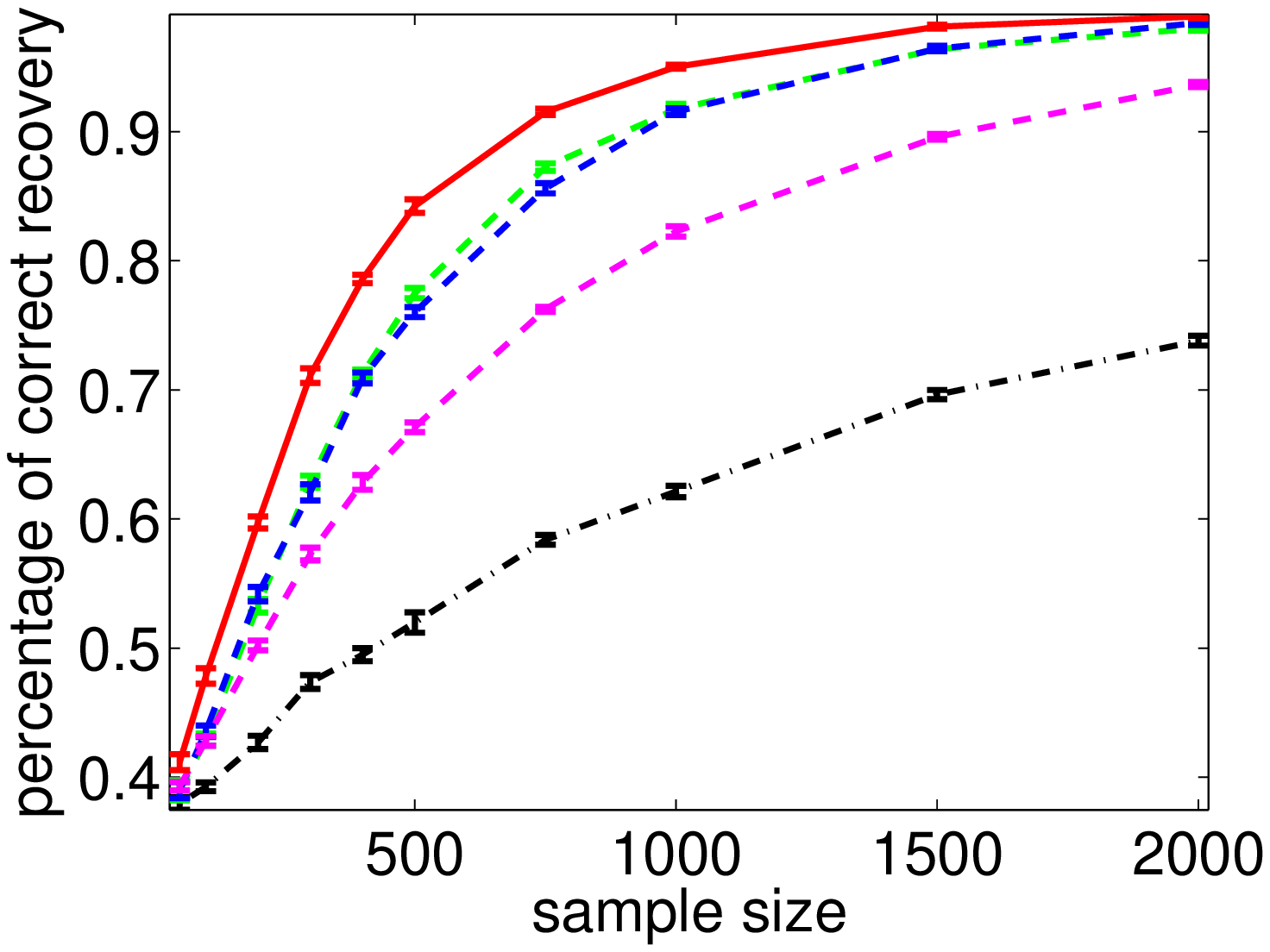}}
  \subfigure[$k=\{4,4\}, \mu=1$]{\label{fig:quartet:f}\includegraphics[width=.25\textwidth]{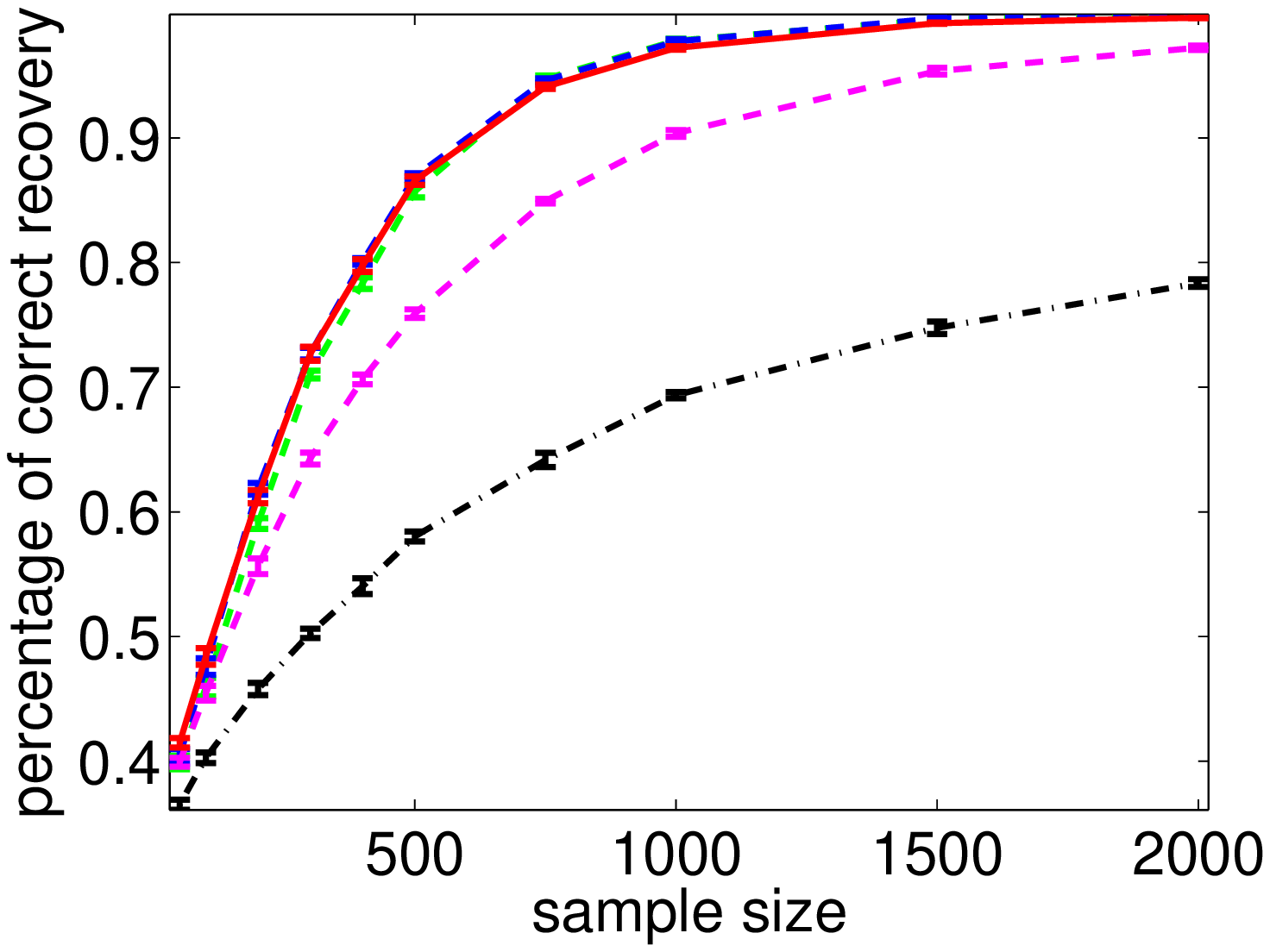}}  
  \\
  \subfigure[$\mu=0.2, \beta=0.5$]{\label{fig:tree:d}\includegraphics[width=.25\textwidth]{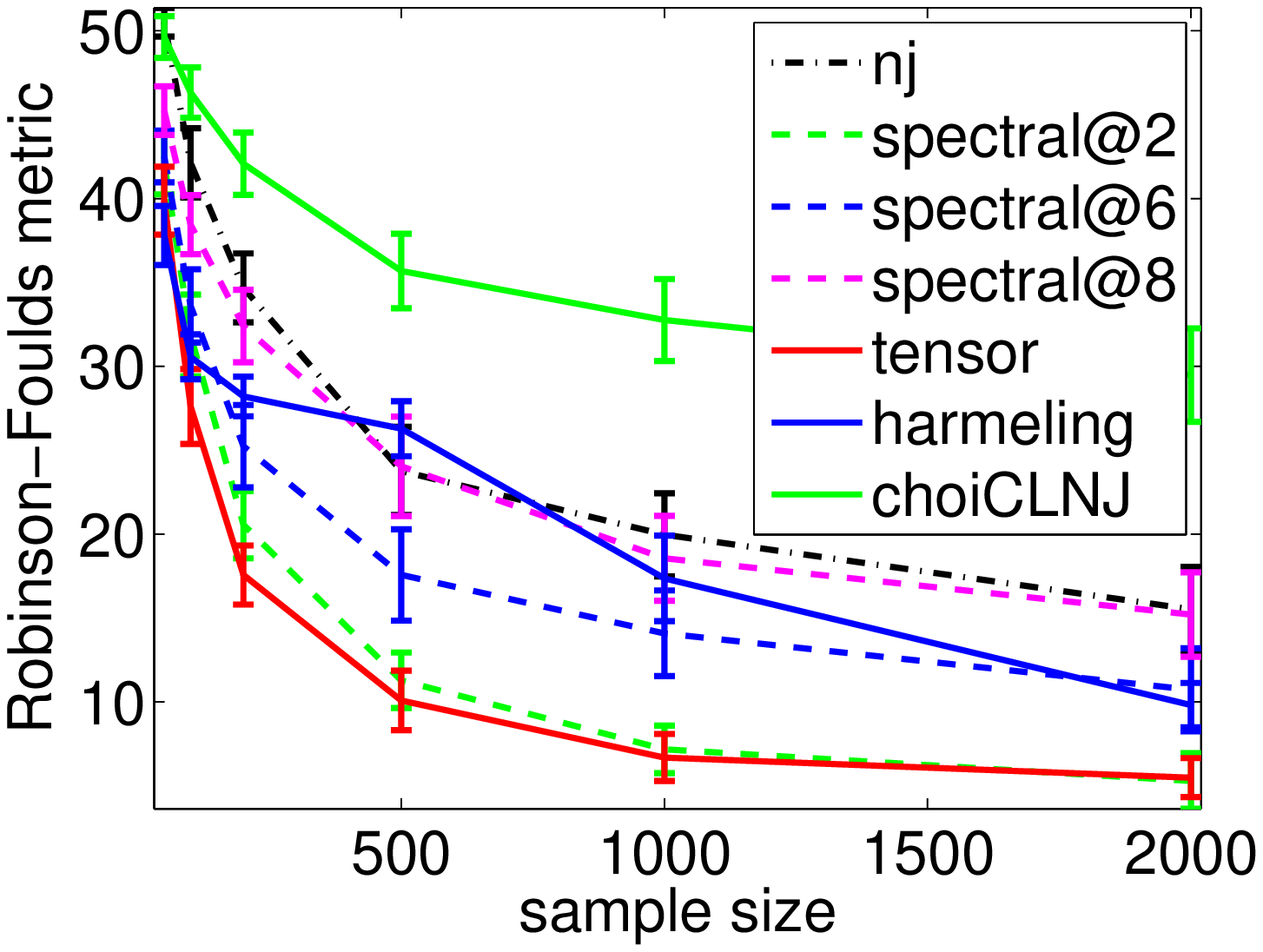}}
  \subfigure[$\mu=0.5, \beta=0.5$]{\label{fig:tree:e}\includegraphics[width=.25\textwidth]{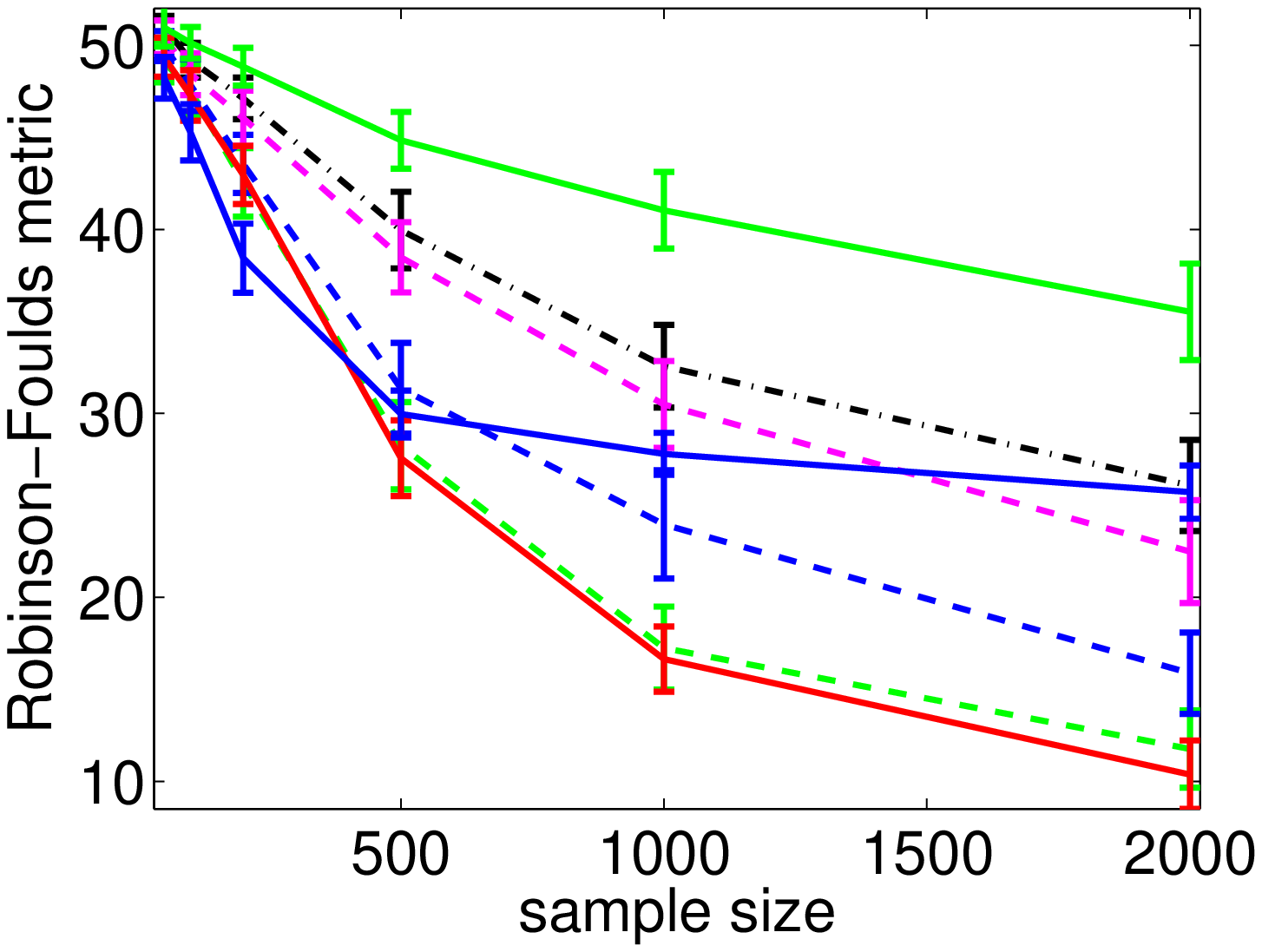}}
  \subfigure[$\mu=1, \beta=0.5$]{\label{fig:tree:f}\includegraphics[width=.25\textwidth]{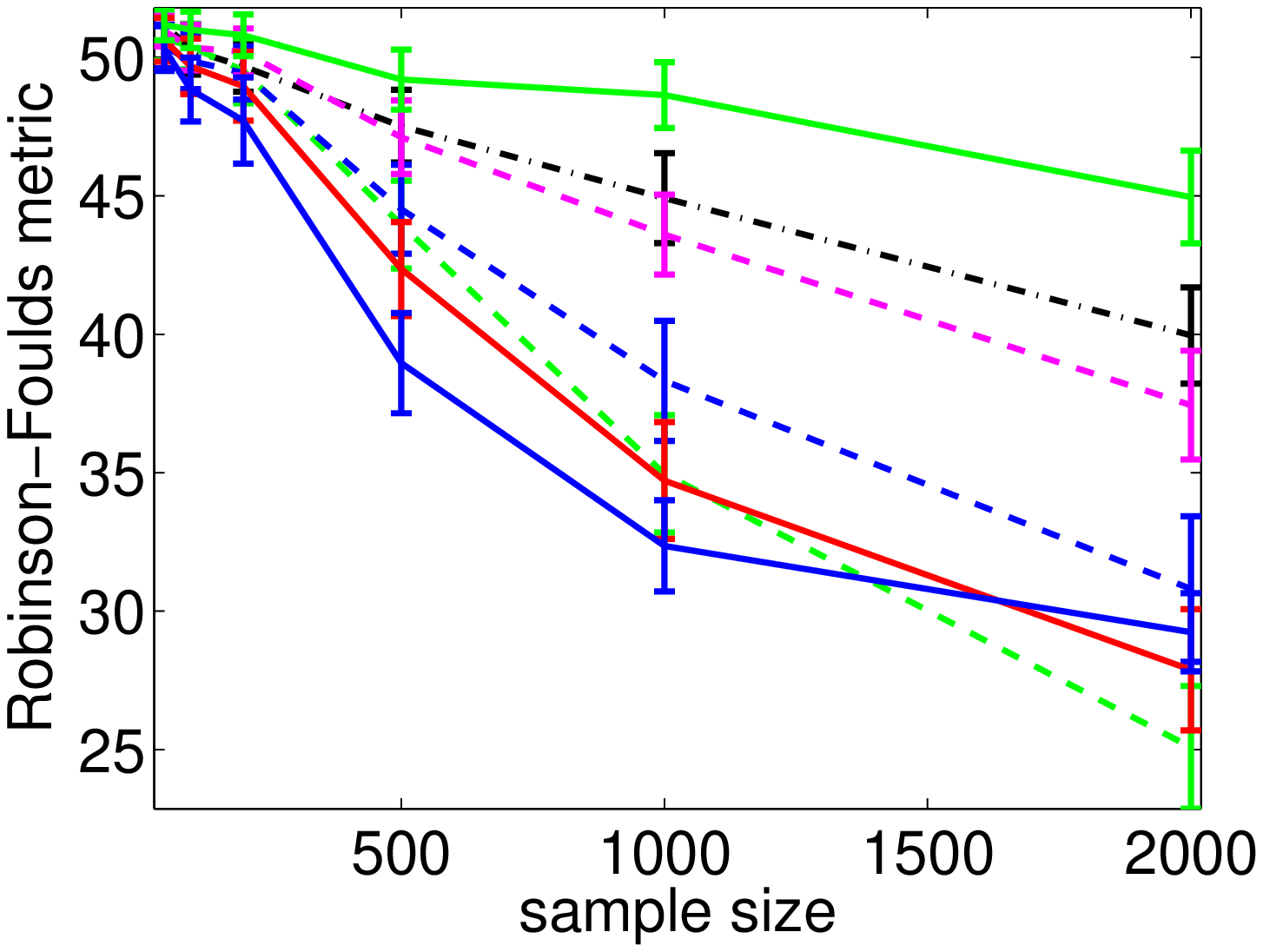}}   \\
  \subfigure[$\mu=0.2, \beta=0.2$]{\label{fig:tree:a}\includegraphics[width=.25\textwidth]{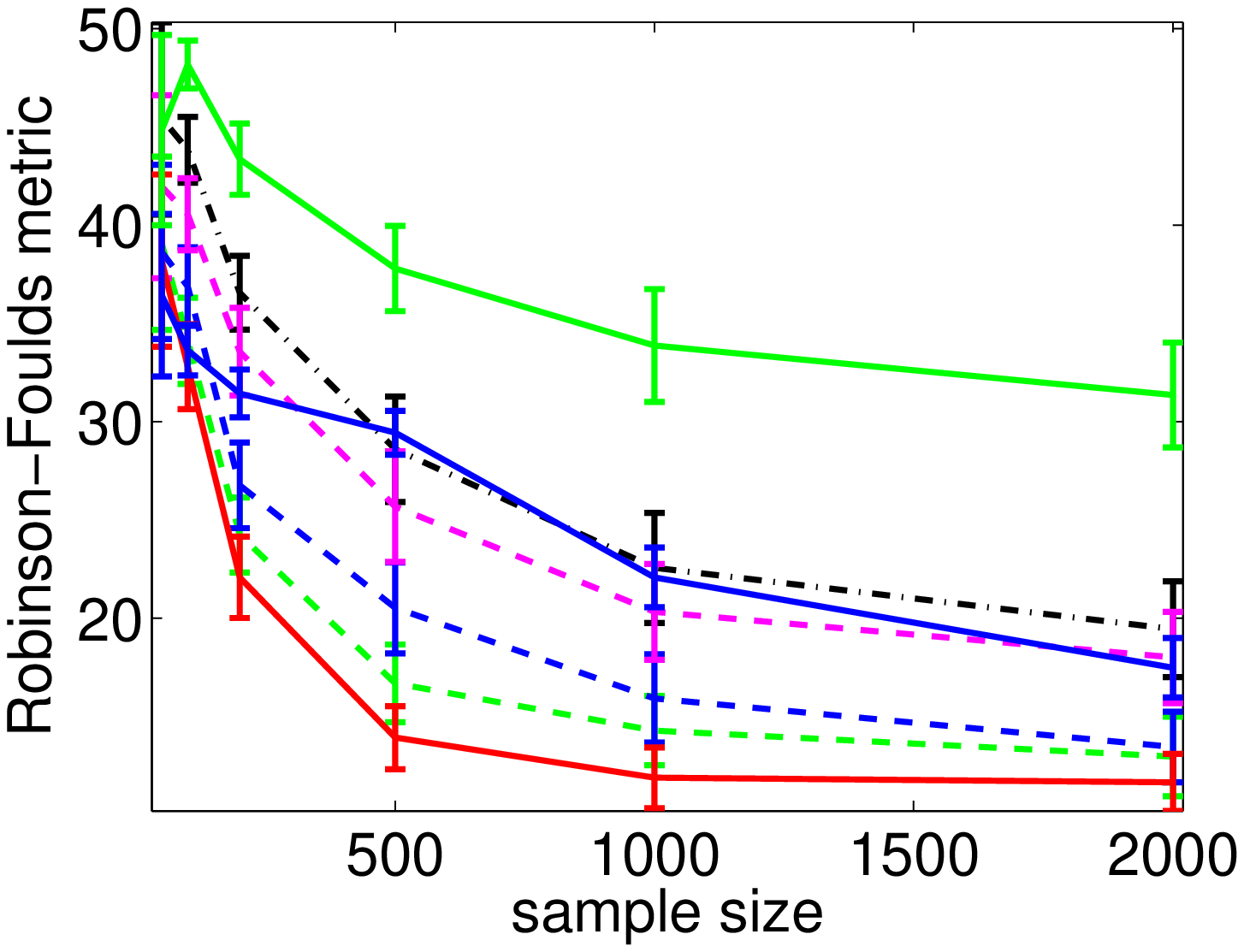}}
  \subfigure[$\mu=0.5, \beta=0.2$]{\label{fig:tree:b}\includegraphics[width=.25\textwidth]{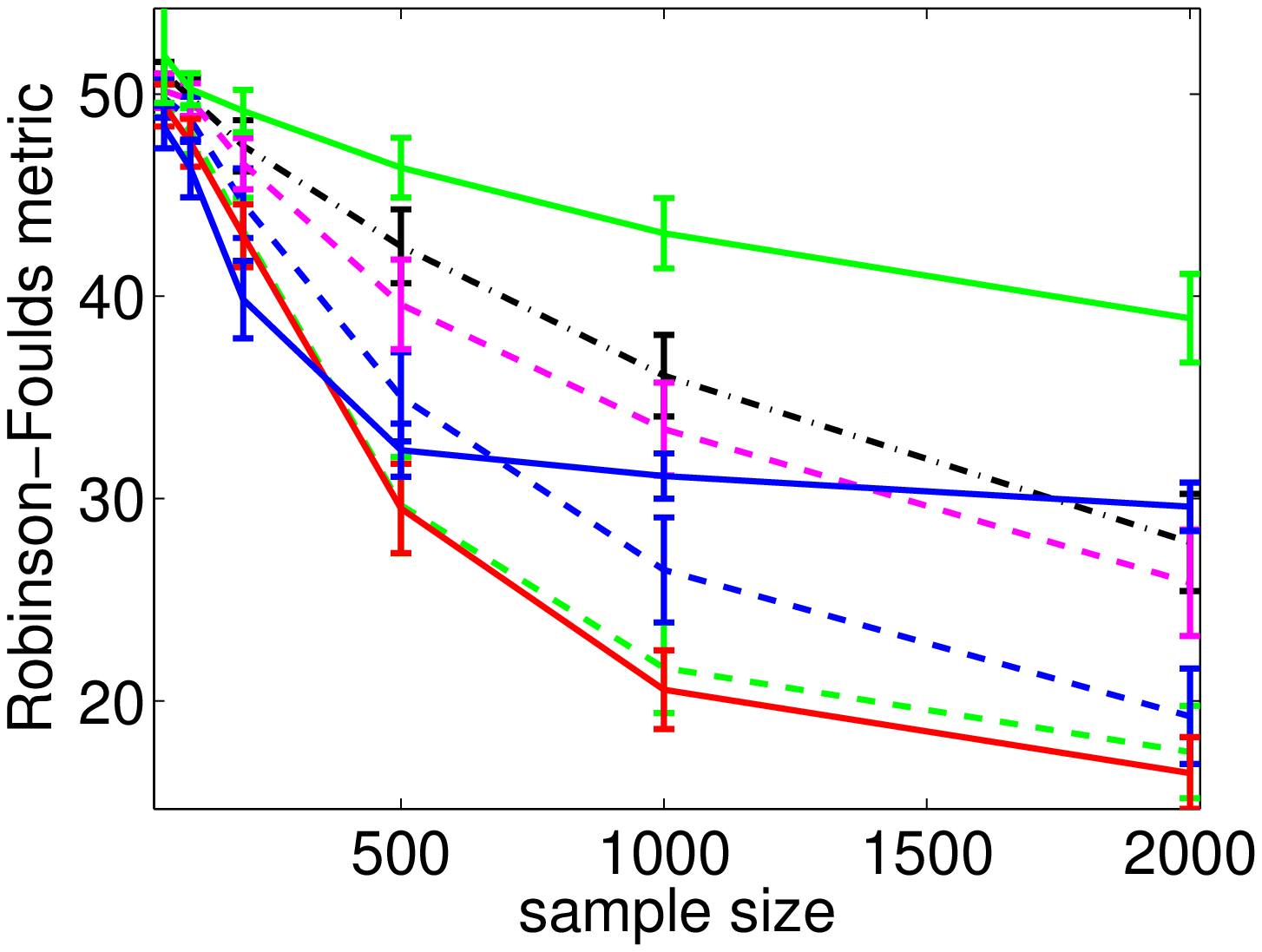}}
  \subfigure[$\mu=1, \beta=0.2$]{\label{fig:tree:c}\includegraphics[width=.25\textwidth]{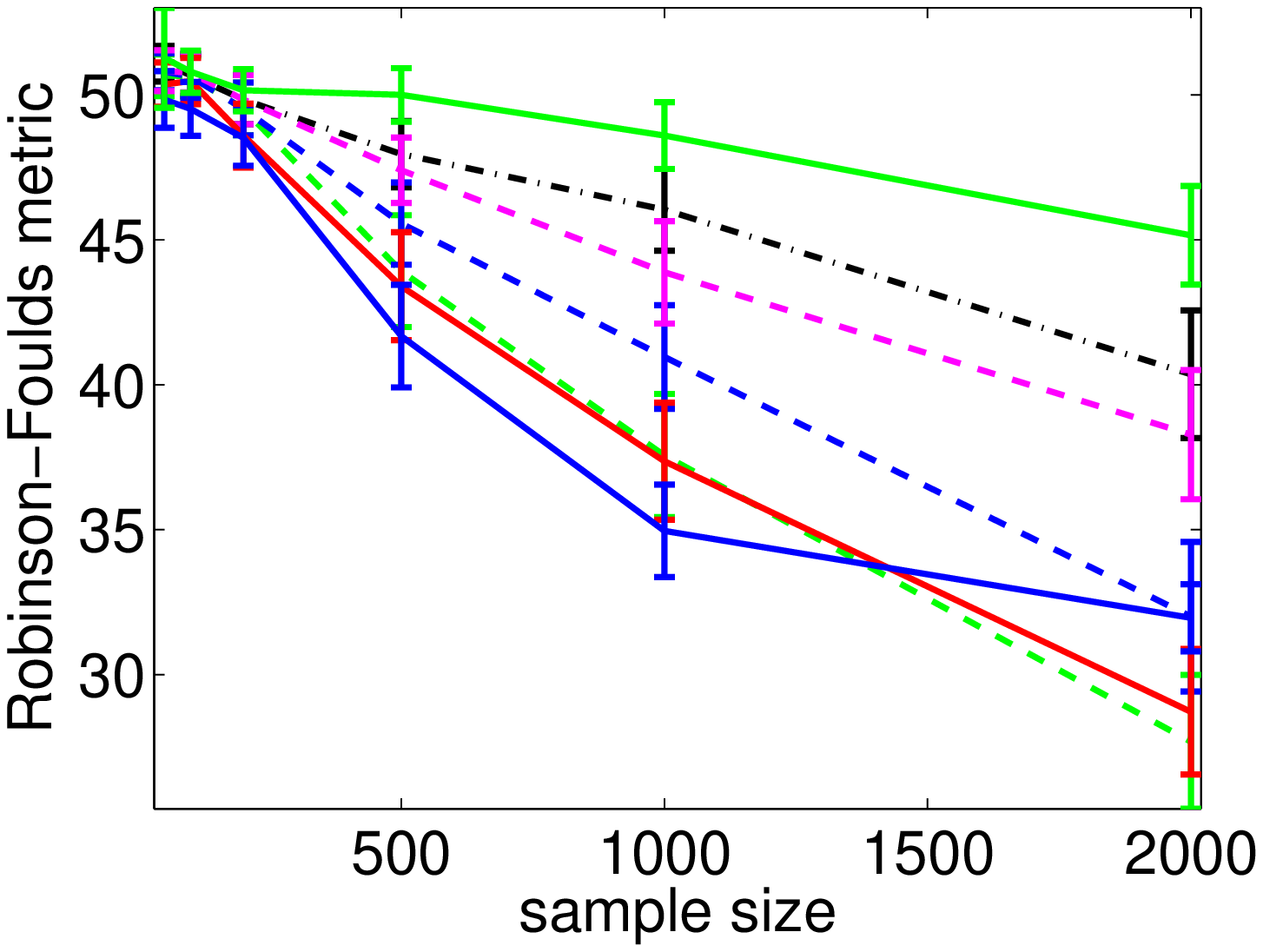}}
  \caption{(a)-(f) Quartet recovery results. (g)-(l) Tree recovery results. ``tensor'' is our method.}
  \label{fig:quartet}
  \vspace{-3mm}
\end{figure*}

We compared our method to NJ and Spectral@$k$ in terms of their ability to recover the quartet relation among four variables. We used quartet with three different configurations for the hidden states: (1) $k_H=2$ and $k_G=4$ (small difference); (2) $k_H=2$, $k_G=8$ (large difference); and (3) $k_H=4$, $k_G=4$ (no difference). In all cases, the states of the observed variables were fixed to $n=10$. In all cases we started from independent $P_{HG}$ but identity $P_{X_i|H}$ and $P_{X_i|G}$, and perturbed them using the following formula 
$
  P(a=i|b) = \frac{P(a=i|b) + u_i}{\sum_i P(a=i|b) + u_i},
$
where all $u_i$ are~\iid~random variables drawn from $\text{Uniform}[0,\mu]$. We then drew random sample from the quartet according to these CPTs. We studied the percentage of correctly recovered quartet relations as we varied the sample size across $S=\{50,$ $100,$ $200,$ $300, 400, 500, 750, 1000, 1500, 2000\}$ and under two different levels of perturbation ($\mu = \{0.5,1\}$). We randomly initialized each experiment 1000 times and report the average quartet recovery performance and the standard error in Figure~\ref{fig:quartet}. 

The proposed method compares favorably to NJ and Spectral@$k$. The performance of Spectral@$k$ varies a lot depending on the chosen number of singular values $k$. Our method is free from tuning parameters and often stays among the top performing ones. Especially when the number of hidden states are very different from each other ($k_H=2$ and $k_G=8$), our method is leading the second best by a large gap~(Figure~\ref{fig:quartet:b} and~\ref{fig:quartet:e}). When both hidden states are the same ($k_H=k_G=4$), the Spectral@$k$ achieves the best performance when the chosen number of singular values $k$ is the same as $k_H$. Note that allowing Spectral@$k$ to use different $k$ resembles using cross validations for finding the best $k$. It is expensive while our approach performs almost indistinguishable from Spectral@$k$ even it choose the best $k$. 

\subsection{Discovering Latent Tree Structure}

We used different tree topologies and sample sizes in this experiment. We generated tree topologies by randomly splitting 16 observed variables recursively into two groups. The recursive splitting stops when there are only two nodes left in a group. We introduced a hidden variable to join the two partitions in each recursion and this gives a latent tree structure. The topology of the tree is controlled by a single splitting parameter $\beta$ which controls the relative size of the first partition versus the second. If $\beta$ is close to $0$ or $1$, we obtain trees of skewed shape, with long path of hidden variables. If $\beta$ is close to $0.5$, the resulting latent trees are more balanced. In our experiments, we experimented with skewed latent trees $\beta=0.2$ and balanced trees $\beta = 0.5$. We first generate different random $k$ between $2$ and $8$ for the hidden states, and then generate the probability models for each tree using the same scheme as in our previous experiment. Here we experimented with perturbation level $\mu=\{0.2, 0.5, 1\}$.

We varied the sample size across $S=\{50, 100, 200, 500,$ $1000,$ $2000\}$, and measured the error of the constructed tree using Robinson-Foulds metric~\citep{RobFou1981}. This measure is a metric over trees of the same number of leaves. It is defined as $(a + b)$ where $a$ is the number of partitions of variables implied by the learned tree but not by the true tree and $b$ is the number of partitions of the variables implied by the true tree but not by the learned tree (in a sense similar to precision and recall score).

The tree recovery results are shown in Figure~\ref{fig:tree:d}-\ref{fig:tree:c}. Again we can see that our proposed method compares favorably to existing algorithms. All through the 6 experimental conditions, the tensor approach and spectral@2 performed the best with sufficiently large sample sizes. Note that we tried out different $k$ for Spectral@$k$ which resembles using cross validations for finding the best $k$. Even in this case, our approach works comparably without having to know $k$.
Harmeling-William's algorithm performed well in small sample sizes, while CLNJ does not perform well in these experimental conditions.

\subsection{Understanding Latent Relations between Stocks}
\begin{figure*}[htb]
  \centering
  \includegraphics[width=1\textwidth]{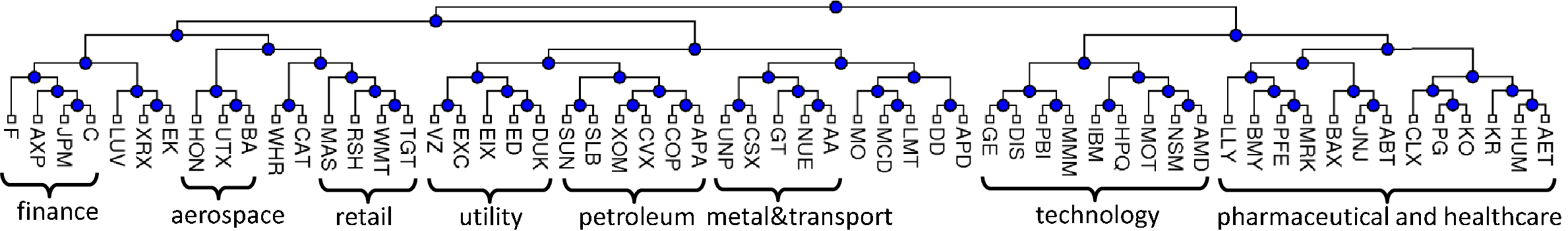}    
  \caption{Latent tree estimated from stock data.}
  \label{fig:stock}
\end{figure*}

We applied our algorithm to discover a latent tree structure from a stock dataset. Our goal is to understand how stock prices $X_i$ are related to each other. We acquired closing prices of 59 stocks from 1984 to 2011 (from www.finance.yahoo.com), which provides us 6800 samples. The daily change of each stock price is discretized into 10 values, and we applied our algorithm to build a latent tree. A visualization of the learned tree topologies and discovered groupings are shown in Figure~\ref{fig:stock}. 


We see nice groupings of stocks according to their industrial sectors. For instance, companies related to petroleum, such as CVX (Chevron), XOM (Exxon Mobil), APA (Apache), COP (ConocoPhillips), SLB (Schlumberger) and SUN (Sunoco), are grouped into a subtree. Pharmaceutical companies, such as MRK (Merck), PFE (Pfizer), BMY (Bristol Myers Squibb), LLY (Eli Lilly), ABT (Abbott Laboratories), JNJ (Johnson and Johnson) and BAX (Baxter International), are all grouped into a subtree. High-tech companies, such as AMD, MOT (Motorola), HPQ (Hewlett-Packard), IBM, are grouped into another subtree. There are also subtree for retailers, such as TGT (Target), WMT (Wal-Mart), RSH (RadioShack), subtree for utility service companies, such as DUK (Duke Energy), ED (Consolidated Edison), EIX (Edison), ECX (Exelon), VZ (Verizon), and subtree related to financial companies, such as C (Citigroup), JPM (JPMorgan Chase), and AXP (American Express).
We can also see subtree related to financial companies, such as C (Citigroup), JPM (JPMorgan Chase), and AXP (American Express). An interesting observation is that F (Ford Motor) which is well-known for its car manufacturing is also placed in the same branch as these financial companies. This seemingly abnormal structure can be explained by the fact that Ford Motor operates under two segments: Automotive and Financial Services. 
Its financial services include the operations of Ford Motor Credit Company and other financial services including holding companies, and real estate. In this respect, it is quite interesting that our algorithm discovered this hidden information.     

We also compared different algorithms in terms of held-out likelihood. We first randomized the data 10 times, and each time used half for training and half for computing the held-out likelihood.  Then we estimated the latent binary tree structures using different algorithms. Finally, we fit latent variable models to the discovered structures. 
The number of the states for all hidden variables, $k$, were the same in each latent variable model. We experimented with $k = 2, 4, 6, 8,10$ to simulate the process of using cross validation to select the best $k$. The results are presented in Table~\ref{tab}. 
\begin{table*}[ht]
\setlength{\extrarowheight}{5pt}
\centering
\caption{Negative log-likelihood ($\times 10^5$) on test data. The small the number the better the method.}
\begin{small}
\begin{tabular}{|c| c| c| c| c| c| c|}
\hline
& Tensor & Spectral@$k$ & Choi (CLNJ) & Neighbor-joining & Harmeling & Chow-Liu\\
\hline
$k = 2$  & $4.41$ &  $4.44$ & $4.43$ & $4.43$ 
  &\multirow{5}{*}{$4.31$} & \multirow{5}{*}{$4.41$}\\
\cline{1-5}
$k = 4$  & $4.30$ & $4.35$ & $4.33$ & $ 4.33$  & &\\
\cline{1-5}
$k = 6$  & $4.28$  & $4.35$ &   $4.32$ & $4.31$  & &\\
\cline{1-5}
$k = 8$  & $\mathbf{4.28}$ & $4.35$ &  $4.32$ & $4.31$  & &\\
\cline{1-5}
$k = 10$  & $4.29$ & $4.37$ &  $4.32$ & $4.31$  & &\\
\hline
\end{tabular}
\end{small}
\label{tab}
\end{table*}
Note that Harmeling-William's algorithm automatically discovers $k$, so it does not use the experimental parameter $k$. 
Chow-Liu tree does not contain any hidden variables and hence just one number in the table. 
CLNJ and Neighbor-joining assume the states for the hidden and observed variables are the same during structure learning. However, in parameter fitting, we can still use different number of hidden states $k$. In this experiment, the structure produced by our tensor approach produced the best held-out likelihood.

\section{Conclusion}
 
In this paper, we propose a quartet-based method for discovering the tree structures of latent variable models. The practical advantage of the new method is that we do not need to pre-specify the number of the hidden states, a quantity usually unknown in practice. The key idea is to view 
the joint probability tables of quadruple of variables as $4$th order tensors and then use the 
spectral properties of the unfolded tensors to design a quartet test. We provide conditions under which the algorithm is consistent and its error probability decays exponentially with increasing the sample size. In both simulated and a real dataset, we demonstrated the usefulness of our methods for discovering latent structures. While in this study we focus on the properties of the $4$th order tensor and its various unfoldings, we believe that properties of tensors and methods and algorithms from multilinear algebra will allow to address many other problems arising from latent variable models.

\clearpage
\newpage
\onecolumn

\begin{center}
{\Large Unfolding Latent Tree Structures using 4th Order Tensors\\[2mm] Appendix}\vspace*{1mm}
\end{center}

\section{Properties and Notations used}
\label{sect:properties}
\underline{Nuclear and Frobenius norms:}
\begin{itemize}
\item Let $\sigma_i$ be the singular values of $A$. Then
  \begin{equation}
  \|A\|_\ast = \sum_i \sigma_i\,,\quad \|A\|_F^2 = \sum_i \sigma_i^2\quad \mbox{ and }\quad \|A\|_F \leq \|A\|_\ast\,.
  \label{def:nuclear_norm}
  \end{equation} 
\item (Nuclear and Frobenius norms are unitarily invariant) For any orthogonal $Q$ we have
  \begin{equation}
  \begin{array}{lclcl}
  \|A\|_\ast & = & \|QA\|_\ast & \!=\! & \|AQ\|_\ast\,,\\[2mm]
  \|A\|_F & = & \|QA\|_F & = & \|AQ\|_F\,.
  \end{array}
  \label{prop:unitarily_invariant}
  \end{equation}
\item $\|AB\|_\ast \leq \|A\|_F\|B\|_F \leq\|A\|_\ast\|B\|_\ast\,$. 
\item Let $\sigma_i$ be the singular values of $X$ and $\tilde\sigma_i$ be the singular values of $\tilde X  = X +E.$
  Then
  \begin{equation}
  \|\mbox{diag}(\tilde\sigma_i - \sigma_i)\|_\ast \leq \|\tilde X - X\|_\ast\,.
  \label{prop:perturb}
  \end{equation}
\end{itemize}\vspace*{3mm}

\noindent\underline{Kronecker and Khatri-Rao products:}
\begin{eqnarray}
\label{prop:Kron_transpose}  (A\otimes B)^\top & = & A^\top \otimes B^\top\\[1mm]
\label{prop:Kron_sum}  (A+B)\otimes C & = & A\otimes C + B \otimes C\\[1mm]
\label{prop:Kron_product}  AB \otimes CD & = & (A\otimes C)(B \otimes D)\\[1mm]
\label{prop:KR_product}  AB \odot CD & = &(A\otimes C)(B \odot D)\\[1mm]
\nonumber  \|A\otimes B\|_F & = & \|A\|_F\|B\|_F\\[1mm]
\nonumber  \mbox{rank}(A\otimes B) & = & \mbox{rank}(A)\, \mbox{rank}(B)
\end{eqnarray}

\underline{Tensor operations:}\\[2mm]
We use the following tensor-matrix products of a tensor ${\cal A} \in\R^{I_1\times I_2\times I_3}$ with matrices
$M^{(n)}\in\R^{J_n\times I_n},\, n= 1,2,3$:
\[\begin{array}{c}
\mbox{\it mode-$1$ product:}\quad({\cal A}\bullet_1 M^{(1)})_{j_1i_2i_3}=
  \displaystyle{\sum\nolimits_{i_1=1}^{I_1} a_{i_1i_2i_3}m^{(1)}_{j_1i_1}\,,}\\[2mm]
\mbox{\it mode-$2$ product:}\quad({\cal A}\bullet_2 M^{(2)})_{i_1j_2i_3}=
  \displaystyle{\sum\nolimits_{i_2=1}^{I_2} a_{i_1i_2i_3}m^{(2)}_{j_2i_2}\,,}\\[2mm]
\mbox{\it mode-$3$ product:}\quad({\cal A}\bullet_3 M^{(3)})_{i_1i_2j_3}=
  \displaystyle{\sum\nolimits_{i_3=1}^{I_3} a_{i_1i_2i_3}m^{(3)}_{j_3i_3}\,,}
\end{array}\]
where $1\leq i_n\leq I_n,\,1\leq j_n\leq J_n.$ These products can be considered as a generalization
of the left and right multiplication of a matrix $A$ with a matrix $M.$ 
The mode-$1$ product signifies multiplying the columns (mode-$1$ vectors) of ${\cal A}$ with the rows of $M^{(1)}$ and similarly for the other tensor-matrix products. 

The {\it contracted product} ${\cal C}$ of two tensors ${\cal A}\in\R^{I\times J \times M}$ and
${\cal B}\in\R^{K\times L \times M}$  along their third modes is a $4$th order tensor denoted by ${\cal C}=\langle {\cal A}, {\cal B} \rangle_3$. ${\cal C}\in \R^{I\times J \times K \times L}$ and its entries ${\cal C}(i,j,k,l),$ $1\leq i \leq I;\, 1 \leq j \leq J;\, 1\leq k \leq K;\, 1 \leq l \leq L$ are defined as $$\quad {\cal C}(i,j,k,l)= \sum\nolimits_{m=1}^M a_{ijm}\;b_{klm}.$$
It can be interpreted as taking inner products of the mode-$3$ vectors of ${\cal A}$ and ${\cal B}$ and storing the results in ${\cal C}$.

The 3 different {\it reshapings}  $A,\,B$ and $C$ (\ref{def:A})--(\ref{def:C}) of the tensor ${\Pcal}$ contain exactly the same entires as $\Pcal$ but in different order.
\begin{tight_list}
  \item $A$ corresponds to the grouping $\{\{1,2\},\{3,4\}\}$ of the variables. The rows of $A$ correspond to dimensions $1$ and $2$ of $\Pcal$, and its columns to dimensions $3$ and $4$. Suppose all observed variables take values from $\{1,\ldots,n\}$, then entry of $A$ at $(x_1+n(x_2-1))$-th row and $(x_3+n(x_4-1))$-th column is equal to $\Pcal(x_1,x_2,x_3,x_4)$; 
  \item $B$ corresponds to the grouping $\{\{1,3\},\{2,4\}\}$, and its entry at $(x_1+n(x_3-1))$-th row and $(x_2+n(x_4-1))$-th column is equal to $\Pcal(x_1,x_2,x_3,x_4)$; 
  \item $C$ corresponds to the grouping $\{\{1,4\},\{2,3\}\}$, and its entry at $(x_1+n(x_4-1))$-th row and $(x_2+n(x_3-1))$-th column is equal to $\Pcal(x_1,x_2,x_3,x_4)$. 
\end{tight_list}

\section{Matrix Representations $A,\,B,\,C$ of $\,{\Pcal}$}
\label{sect:fromP_toABC}

\underline{From ${\cal P}$ to $A,\,B,\,C$:}\\[2mm]
Let $X\in\R^{m\times k},\,Y\in \R^{k\times l}$, $Z\in\R^{n\times l},$
$X=(x_1,\ldots, x_k)$ and $Z=(z_1,\ldots, z_l)$. A useful property that we will use in our derivations is the following
\begin{equation}
X\,Y\,Z^\top = \sum_{i,j} x_i \, y_{ij}\, z_j^\top.
\label{matrix_multiplication}
\end{equation}

We can derive the formula for $A$ starting from the element-wise formula (\ref{def:P})
$$ \Pcal (x_1, x_2, x_3, x_4) = \sum_{h, g} P(x_1 | h) P(x_2 | h) P(h,g) P(x_3|g) P(x_4 | g)$$
and placing all entries in the matrix $A$ in the correct order.
Note that given $h$ and $g$ we only need one column of each $P_{1|H},\,P_{2|H},\,P_{3|G}$ and $P_{4|G},$
which we will denote by $(P_{1|H})_h,\,(P_{2|H})_h,\,(P_{3|G})_g$ and $(P_{4|G})_g.$
In order to obtain a matrix such that $X_1$ and $X_2$ are mapped to rows and  $X_3$ and $X_4$
are mapped to columns, 
we need to map all possible products of single element of $(P_{1|H})_h$ and single element of $(P_{2|H})_h$ to rows and 
and similarly, we need to map all possible products of single element of $(P_{3|G})_g$ and single element of
$(P_{4|G})_g$ to columns.
This can be done using Khatri-Rao products in the following way
\begin{equation*}
\begin{array}{rcl}
A 
& = & \displaystyle{\sum_{h,g}  \Big((P_{2|H})_h\odot (P_{1|H})_h\Big) \,\,(P_{HG})_{hg} \,\,
  \Big((P_{4|G})_g\odot (P_{3|G})_g\Big)^\top}\\[3mm]
& \stackrel{(\ref{matrix_multiplication})}{=} & \big(P_{2|H} \odot P_{1|H}\big)\,\,\, P_{HG} \,\,\, \big(P_{4|G} \odot P_{3|G}\big)^\top.
\end{array}
\end{equation*}

The matrix $B$ is unfolding of ${\cal P}$,
such that the rows of $B$ correspond to $X_1$ and $X_3$ and the columns of
$B$ correspond to $X_2$ and $X_4.$
We have
\begin{equation*}
\begin{array}{rcl}
B 
& = & \displaystyle{\sum_{h,g}  \Big((P_{3|G})_g\odot (P_{1|H})_h\Big) \,\,(P_{HG})_{hg} \,\,
  \Big((P_{4|G})_g\odot (P_{2|H})_h\Big)^\top}\\[3mm]
& \stackrel{(\ref{prop:Kron_transpose})}{=}  & \displaystyle{\sum_{h,g}  \Big((P_{3|G})_g\otimes (P_{1|H})_h\Big) \,\,(P_{HG})_{hg} \,\,
  \Big((P_{4|G})_g^\top\otimes (P_{2|H})_h^\top\Big)}\\[3mm]
&\stackrel{(\ref{prop:Kron_product})}{=} & \displaystyle{\sum_{h,g} (P_{HG})_{hg} \,\,\Big((P_{3|G})_g(P_{4|G})_g^\top\Big) 
  \otimes \Big((P_{1|H})_h (P_{2|H})_h^\top\Big)}\\[3mm]
&\stackrel{(\ref{prop:Kron_sum})}{=} &  \displaystyle{\sum_{h} \Big(\sum_g (P_{HG})_{hg} (P_{3|G})_g(P_{4|G})_g^\top\Big) 
  \otimes \Big((P_{1|H})_h (P_{2|H})_h^\top\Big)}\\[3mm]
&\stackrel{(\ref{matrix_multiplication})}{=} & \displaystyle{\sum_{h} \Big(P_{3|G}\,\,\mbox{diag}((P_{HG})_h)\, P_{4|G}^\top \Big)
  \otimes \Big((P_{1|H})_h (P_{2|H})_h^\top\Big)}\\[3mm]
&\stackrel{(\ref{prop:Kron_product})}{=}  & \displaystyle{\sum_{h} \Big(P_{3|G}\otimes (P_{1|H})_h\Big)\,\, \mbox{diag}((P_{HG})_h)\,\,
   \Big(P_{4|G}^\top \otimes (P_{2|H})_h^\top\Big)}\\[3mm]
&\stackrel{\mbox{\scriptsize block}-(\ref{matrix_multiplication})}{=}& \big(P_{3|G}\otimes P_{1|H}\big) \,\,\mbox{diag}(P_{HG}(:)) \,\, \big(P_{4|G}^\top \otimes P_{2|H}^\top\big)\\[2mm]
&\stackrel{(\ref{prop:Kron_transpose})}{=} & \big(P_{3|G} \otimes P_{1|H}\big)\,\,\, \mbox{diag}(P_{HG}(:))\,\,\, \big(P_{4|G} \otimes P_{2|H}\big)^\top.
\end{array}
\end{equation*}
The expression for $C$ is derived in a similar way.

\underline{Other representations of $A,\,B,\,C$:}\\[2mm]
Using the properties in Section~\ref{sect:properties} and the formulas (\ref{eq:Acompact})--(\ref{eq:Ccompact})
for the matrix unfoldings $A,\,B$ and $C$, we can derive the following additional 
formulas,
\begin{equation}\begin{array}{rcl}
A & = &   \quad \big(P_{2|H} \odot P_{1|H}\big)\,\,\, P_{HG} \,\,\, \big(P_{4|G} \odot P_{3|G}\big)^\top\\[1mm]
& = &   \quad \big(I_n\, P_{2|H} \odot P_{1|H}\,I_H \big)\,\,\, P_{HG} \,\,\, \big(I_n\, P_{4|G} \odot P_{3|G}\,I_G\big)^\top\\[1mm]
& \stackrel{(\ref{prop:KR_product})}{=} & \quad\big(I_{n} \otimes P_{1|H}\!\big)\quad \big (P_{2|H} \odot I_{H}\!\big)\quad P_{HG}\quad
  \big(P_{4|G} \odot I_{G}\big)^\top\quad \big( I_{n} \otimes P_{3|G}\big)^\top\\[2mm]
& = &
        \!\!\left(\begin{array}{l}
          \hspace*{-2mm}P_{1|H}\\ [5mm]
          \hspace*{.7cm} \ddots \\[5mm]
          \hspace*{12mm} P_{1|H} \hspace*{-2mm}
        \end{array}\right)
        \left(\begin{array}{l}
          \hspace*{-2mm}p_{2|H}^{(1,1)}\\
          \hspace*{2mm}p_{2|H}^{(1,2)}\\[-2mm]
          \hspace*{8mm}\ddots\hspace*{-2mm}\\
          \hspace*{-2mm}p_{2|H}^{(2,1)}\\
          \vdots \hspace*{3mm} \ddots
        \end{array}\right)
        \,P_{HG}\,
        \left(\begin{array}{l}
          \hspace*{-2mm}p_{4|G}^{(1,1)}\\
          \hspace*{2mm}p_{4|G}^{(1,2)}\\[-2mm]
          \hspace*{8mm}\ddots\hspace*{-2mm}\\
          \hspace*{-2mm}p_{4|G}^{(2,1)}\\
          \vdots \hspace*{3mm} \ddots
        \end{array}\right)^\top
        \!\left(\begin{array}{l}
          \hspace*{-2mm}P_{3|G}\\ [5mm]
          \hspace*{.7cm} \ddots \\[5mm]
          \hspace*{12mm} P_{3|G} \hspace*{-2mm}
        \end{array}\right)^\top\!\!\!,  
\end{array}
\label{eq:A}
\end{equation}
\begin{equation}\begin{array}{rcl}
B & = & \quad\quad \big(P_{3|G} \otimes P_{1|H}\big)\,\,\, \mbox{diag}(P_{HG}(:))\,\,\, \big(P_{4|G} \otimes P_{2|H}\big)^\top\\[1mm]
& = & \quad\quad \big(P_{3|G} \, I_G \otimes I_n\,P_{1|H}\big)\,\,\, \mbox{diag}(P_{HG}(:))\,\,\, 
  \big(P_{4|G}\, I_G \otimes I_n\,P_{2|H}\big)^\top\\[1mm]
& \stackrel{(\ref{prop:Kron_product}),(\ref{prop:Kron_transpose})}{=} & 
  \quad\quad\Big(P_{3|G}  \otimes I_{n}\!\Big)\quad\, \Big(I_{G} \otimes P_{1|H}\!\Big)\,\,
  \quad \mbox{diag}(P_{HG}(:))\,\quad
   \Big(I_{G} \otimes P_{2|H}\!\Big)^\top\quad\, \Big(P_{4|G} \otimes I_{n}\!\Big)^\top\\[2mm]
& = &
        \left(\begin{array}{cc}
          \!\!\!\big(p_{3|G}^{(1,1)}\big)  & \cdots \\[5mm]
          \!\!\!\big(p_{3|G}^{(2,1)}\big)  & \\ [2mm]
          \hspace*{4mm}\vdots &
        \end{array}\right)
        \left(\begin{array}{l}
          \hspace*{-2mm}P_{1|H}\\ [5mm]
          \hspace*{.7cm} \ddots \\[5mm]
          \hspace*{12mm} P_{1|H} \hspace*{-2mm}
        \end{array}\right)
        \mbox{diag}(P_{HG}(:))
        \left(\begin{array}{l}
          \hspace*{-2mm}P_{2|H}\\ [5mm]
          \hspace*{.7cm} \ddots \\[5mm]
          \hspace*{12mm} P_{2|H} \hspace*{-2mm}
        \end{array}\right)^\top
        \!\!\!\left(\begin{array}{cc}
          \!\!\!\big(p_{4|G}^{(1,1)}\big) & \cdots \\[5mm]
          \!\!\!\big(p_{4|G}^{(2,1)}\big)  & \\ [2mm]
          \hspace*{4mm}\vdots &
        \end{array}\right)^\top\!\!,    
\end{array}
\label{eq:B}
\end{equation}
where $(p^{(i,j)})$ is a diagonal block of size ($n \times n$) with all diagonal elements equal to $p^{(i,j)}.$

The formula for $C$ can be obtained from the ones for B by swapping the positions of $P_{3|G}$ and $P_{4|G}.$\\

\underline{Rank properties of $A,\,B,\,C$:}\\[2mm]
In this section we prove the rank properties used in Section~\ref{sect:rank_properties} of the paper.

{\bf Lemma.} {\it If $X\in \R^{m\times n},\,Y\in \R^{n\times k},\,Z\in \R^{l\times m}$,
$Y$ has full row rank, and $Z$ has full column rank, then
$$\textnormal{rank}(XY) = \textnormal{rank}(X),$$
$$\textnormal{rank}(ZX) = \textnormal{rank}(X).$$}

We assume that all CPTs have full column (or row) rank. Then the first two matrices in (\ref{eq:A})
also have full column rank. 
The last two matrices have full row rank. From the lemma, it follows that
\begin{equation}
\mbox{rank}(A) = \mbox{rank}(P_{HG}) = k
\label{eq:rankA}
\end{equation}

Analogously, the first two matrices in (\ref{eq:B}) have full column rank. 
The last two matrices have full row rank. From the lemma, it follows that
\begin{equation}
\mbox{rank}(B) = \mbox{nnz}(P_{HG}),
\label{eq:rankB}
\end{equation}
i.e., generically, 
$$\mbox{rank}(B) = k^2.$$

\section{Algorithms}
\label{app:build_tree}
$ $\vspace*{-3mm}
\begin{algorithm}[h!]
\caption{$\Tcal_{next}=$ QuartetTree($\Tcal_1$, $\Tcal_2$, $\Tcal_3$, $X_4$)}
\begin{algorithmic}[1] 
  \REQUIRE Leaf($\Tcal$): leaves of a tree $\Tcal$;  
  \FOR{$j=1$ to $3$}    
    \STATE $X_i \leftarrow$ Randomly choose a variable from Leaf($\Tcal_i$)
  \ENDFOR
  \STATE $i^\ast \leftarrow$ Quartet($X_1$, $X_2$, $X_3$, $X_4$),~~~~$\Tcal_{next} \leftarrow \Tcal_{i^\ast}$
\end{algorithmic} 
\end{algorithm}%
\begin{algorithm}[h!]
\caption{$\Tcal=$ Insert$(\Tcal, \widetilde{\Tcal}, X_i)$}
\label{alg:insert}
\begin{algorithmic}[1]
\REQUIRE Left($\Tcal$) and Right($\Tcal$): left and right child branch of the root respectively; $\Tcal + \Tcal'$: return a new tree connecting the root of two trees by an edge and use the root of $\Tcal$ as the new root
\IF{$|$Leaf$(\Tcal)|=1$} 
  \STATE $\Tcal$ $\leftarrow$ Form a tree with root $R$ connecting Leaf$(\Tcal)$ and $X_i$.
\ELSE
  \STATE $\Tcal_{next} \leftarrow$ QuartetTree(Left($\Tcal$), Right($\Tcal$), $\widetilde{\Tcal}$, $X_i$)
  \IF{$\Tcal_{next}=$ Left($\Tcal$)}
      \STATE $\Tcal \leftarrow$ Insert($\Tcal_{next}$, Right($\Tcal$) $+$ $\widetilde{\Tcal}$, $X_i$)
  \ELSIF{$\Tcal_{next}=$ Right($\Tcal$)}
      \STATE $\Tcal \leftarrow$ Insert($\Tcal_{next}$, Left($\Tcal$) $+$ $\widetilde{\Tcal}$, $X_i$)
  \ENDIF  
\ENDIF
\STATE $\Tcal \leftarrow \Tcal + \widetilde{\Tcal}$
\end{algorithmic} 
\label{alg:quartettree}
\end{algorithm}
\begin{algorithm}[h!]
\caption{$\Tcal=$ BuildTree($\{X_1,\ldots,X_d\}$)}
\begin{algorithmic}[1]
  \STATE Randomly choose $X_1$, $X_2$, $X_3$ and $X_4$
  \STATE $i^\ast \leftarrow$ Quartet($X_1$, $X_2$, $X_3$, $X_4$)  
  \STATE $\Tcal \leftarrow$ Form a tree with two connecting hidden variables $H$ and $G$, where $H$ joins $X_{i^\ast}$ and $X_4$, while $G$ joins variables in $\{X_1,X_2,X_3\}\setminus \{X_{i^\ast}\}$
  \FOR{$i=5~\text{to}~d$}
    \STATE Pick a root $R$ from $\Tcal$ which split it to three branches of equal sizes, and $\Tcal_{next} \leftarrow$ QuartetTree(Left($\Tcal$), Right($\Tcal$), Middle($\Tcal$), $X_i$)
    \IF{$\Tcal_{next}=$ Left($\Tcal$)}
        \STATE $\Tcal \leftarrow$ Insert($\Tcal_{next}$, Right($\Tcal$) $+$ Middle($\Tcal$), $X_i$)
    \ELSIF{$\Tcal_{next}=$ Right($\Tcal$)}        
        \STATE $\Tcal \leftarrow$ Insert($\Tcal_{next}$, Left($\Tcal$) $+$ Middle($\Tcal$), $X_i$)
    \ELSIF{$\Tcal_{next}=$ Middle($\Tcal$)}
        \STATE $\Tcal \leftarrow$ Insert($\Tcal_{next}$, Right($\Tcal$) $+$ Left($\Tcal$), $X_i$)
    \ENDIF  
  \ENDFOR
\end{algorithmic} 
\label{alg:buildtree_extended}
\end{algorithm}

\section{Recovery Conditions for Quartet}
\label{sect:perturbation}

{\bf Latent variables $H$ and $G$ are independent.} In this case, rank($P_{HG})=1$, since $P(h,g)=P(h)P(g)$. Applying the relation in Equation~\ref{eq:rankA_}, we have that $rank(A)=1 \ll rank(B)$. Furthermore, since $A$ has only one nonzero singular value, we have $\|A\|_\ast = \|A\|_F = \|B\|_F \leq \|B\|_\ast$, since $\|M\|_F \leq \|M\|_\ast$ for any $M.$
In this case, we know for sure that the nuclear norm quartet test will return the correct topology. 

{\bf Latent variables $H$ and $G$ are not independent.} We analyze this case by treating it as perturbation $\Delta$ away from the $P_{HG}$ in the independent case. We want to characterize how large $\Delta$ can be while still allowing the nuclear norm quartet test to find the correct latent relation. 
Suppose $A_{\perp}$ and $B_{\perp}$ are the unfolding matrices in the case where $H$ and $G$ are independent. Suppose we add perturbation $\Delta$ to $P_{HG}$, then $A_{\perp} = \big(P_{2|H} \odot P_{1|H}\big)\,\,\, P_{HG} \,\,\, \big(P_{4|G} \odot P_{3|G}\big)^\top$ and its perturbed version is $A = \big(P_{2|H} \odot P_{1|H}\big)\,\,\, (P_{HG} + \Delta) \,\,\, \big(P_{4|G} \odot P_{3|G}\big)^\top$. We want to bound the difference $\abr{\,\|A_\perp\|_{\ast} - \|A\|_{\ast}}$.
We have 
\begin{align*}
\hspace*{-12mm}\abr{\,\|A_\perp\|_{\ast} - \|A\|_{\ast}} = &\, \Big|{\sum\nolimits_i \sigma_i(A_\perp) - \sum_i \sigma_i(A)}\Big| \nonumber \\ 
  \leq &\, \sum\nolimits_i \abr{\sigma_i(A_\perp) - \sigma_i(A)} \\
  \stackrel{(\ref{prop:perturb})}{\leq} &\, \nbr{A_\perp - A}_{\ast} \\
  \leq &\, \big \| \big(P_{2|H} \odot P_{1|H}\big)\,\,\, \Delta \,\,\, \big(P_{4|G} \odot P_{3|G}\big)^\top \big \|_{\ast} \nonumber \\[2mm]
  \leq &\, \big \| P_{2|H} \odot P_{1|H}\big \|_F\,\,\, \|\Delta\|_F \,\,\, \big \|P_{4|G} \odot P_{3|G} \big \|_{F} \nonumber \\[2mm]
  \leq &\,   k\nbr{\Delta}_{F}, \nonumber
\end{align*}
since $P_{2|H} \odot P_{1|H}$ and $P_{4|G} \odot P_{3|G}$ are CPTs with $k$ columns each,
and thus $\big \| P_{2|H} \odot P_{1|H}\big \|_F^2 \leq k$
and $\big \|P_{4|G} \odot P_{3|G} \big \|_{F}^2 \leq k$.

Analogously, $B_{\perp} = \big(P_{3|G} \otimes P_{1|H}\big)\,\,\, \diag(P_{HG}(:)) \,\,\, \big(P_{4|G} \otimes P_{2|H}\big)^\top$ and its perturbed version is $B = \big(P_{3|G} \otimes P_{1|H}\big)\,\,\, \diag(P_{HG}(:) + \Delta(:)) \,\,\, \big(P_{4|G} \otimes P_{2|H}\big)^\top$. 
We want to bound the difference $\abr{\|B_{\perp}\|_{\ast} - \|B\|_{\ast}}$. 
We have 
\begin{align*}
\abr{\|B_{\perp}\|_{\ast} - \|B\|_{\ast}}  = &\,  \Big|{\sum\nolimits_i \sigma_i(B_{\perp}) - \sum_i \sigma_i(B)}\Big| \nonumber \\  
  \leq &\,  \sum\nolimits_i \abr{\sigma_i(B_{\perp}) - \sigma_i(B)} \\
  \stackrel{(\ref{prop:perturb})}{\leq} &\,  \nbr{B_{\perp} - B}_{\ast} \\    
  \leq &\,  \big \| \big(P_{3|G} \otimes P_{1|H}\big)\,\,\, \diag(\Delta(:)) \,\,\, \big(P_{4|G} \otimes P_{2|H}\big)^\top \big \|_{\ast} \nonumber \\[2mm] 
  \leq &\,  \big \| P_{3|G} \otimes P_{1|H}\big \|_F \,\,\, \big \|\diag(\Delta(:))\big \|_F \,\,\, \big \|P_{4|G} \otimes P_{2|H} \big \|_F \nonumber \\[2mm] 
  \leq &\,   k^2 \nbr{\diag(\Delta(:))}_F\\[2mm] 
  = &\,  k^2 \nbr{\Delta}_F, 
\end{align*}
since $P_{3|G} \otimes P_{1|H}$ and $P_{4|G} \otimes P_{2|H}$ are CPTs with $k^2$ 
columns, and thus $\big \| P_{3|G} \otimes P_{1|H}\big \|_F^2 \leq k^2$
and $\big \|P_{4|G} \otimes P_{2|H} \big \|_F^2 \leq k^2$.

Therefore, we get the following upper and lower bound:
\begin{align*}
  & \|A\|_{\ast} \leq \|A_\perp\|_{\ast} + \,k\,\nbr{\Delta}_F,\\
  & \|B\|_{\ast} \geq \|B_\perp\|_{\ast} - k^2\nbr{\Delta}_F. 
\end{align*}
If we require that
\begin{align*}
  &\|A_\perp\|_{\ast} + k\nbr{\Delta}_F 
  \leq \|B_\perp\|_{\ast} - k^2\nbr{\Delta}_F, 
\end{align*} 
then we will have $\|A\|_{\ast}\leq \|B\|_{\ast}$. 

We can derive similar condition for the relationship $\|A\|_\ast\leftrightarrow \|C\|_\ast.$
Let $$\theta := \min \{\|B_\perp\|_\ast - \|A_\perp\|_\ast,~\|C_\perp\|_\ast - \|A_\perp\|_\ast\}.$$
We thus obtain an upper bound on the allowed perturbation: 
\begin{align}
\label{cond:delta_quartet}
    \nbr{\Delta}_F 
  \leq \frac{\theta}    {k^2 +k}\,.
\end{align}

\section{Recovery Conditions for Latent Tree}
\label{app:recovery_tree}

When latent variables $H$ and $G$ are independent, we have that $P_{HG} = P_H P_G^\top$. In this case,
\begin{equation}
\begin{array}{rcl}
 \nbr{B_\perp}_\ast
&=& \nbr{(P_{3|G} \otimes P_{1|H}) (\diag(P_G) \otimes \diag(P_H)) (P_{4|G}\otimes P_{2|H})^\top}_\ast  \\[1mm]
&= &\nbr{(P_{3|G} \diag(P_G) P_{4|G}^\top) \otimes (P_{1|H} \diag(P_H) P_{2|H}^\top)}_\ast  \\[1mm]
&= &\nbr{P_{34} \otimes P_{12}}_\ast \\[1mm]
&\geq& \nbr{P_{34} \otimes P_{12}}_F
\end{array}
\label{eq:B_perp_nn}
\end{equation}
and
\begin{equation}
\hspace*{-2cm}\begin{array}{rcl}
\nbr{A_\perp}_\ast
&=& \nbr{ (P_{2|H} \odot P_{1|H})\,\,\, P_H P_G^\top \,\,\, (P_{4|G} \odot P_{3|G})^\top}_\ast  \\[1mm]
&=& \nbr{P_{12}(:) P_{34}(:)^\top}_\ast  \\[1mm]
&=& \nbr{P_{12}(:) P_{34}(:)^\top}_F  \\[1mm]
&=& \nbr{P_{34} \otimes P_{12}}_F
\end{array}
\label{eq:A_perp_nn}
\end{equation}
and thus\\[2mm]
$ \hspace*{3.6cm}\nbr{A_\perp}_\ast \,\, \leq\,\,  \nbr{B_\perp}_\ast\,.$

Suppose now that $H$ and $G$ are not independent and thus we have $P_{HG} = P_H P_G^\top + \Delta.$
The goal is to characterize all $\Delta$s, such that $ \nbr{A}_\ast \leq  \nbr{B}_\ast$ still holds for any quartet.
From the above formulas it follows that the upper bound on $\Delta$ depends only  on 
pairwise marginal distributions.

Since the perturbed version of $P_H P_G^\top$ remains a joint probability table,
all entries of the perturbation matrix $\Delta$ have to sum to 0,~\ie,~$\one^\top \Delta(:)=0$. We further assume that each column sum and each
row sum of $\Delta$ is also equal to $0$,~\ie,~$\one^\top \Delta = \zero$ and $\Delta\, \one = \zero$. In this case, $\one^\top \Delta(:)=0$ is satisfied automatically.

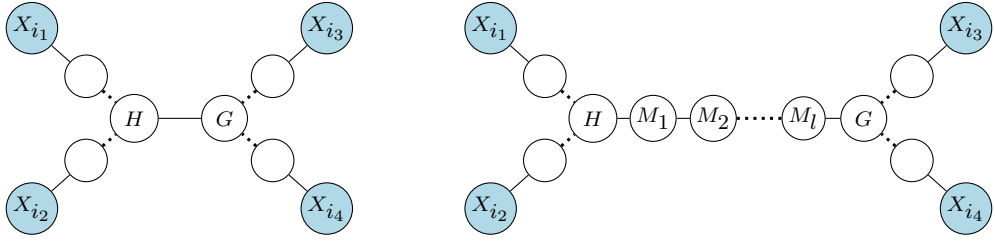
\begin{figure}[ht]
    \renewcommand{\arraystretch}{1}     
    \setlength{\tabcolsep}{5pt} 
    \begin{tabular}{cc}
    \begin{tikzpicture}
    [
      scale=0.8,
      observed/.style={circle,inner sep=0.3mm,draw=black,fill=MyBlue}, 
      hidden/.style={circle,inner sep=1.1mm,draw=black},
      hidden2/.style={circle,inner sep=0.3mm,draw=black},
      hidden3/.style={circle,inner sep=2mm,draw=black},
    ]
    \node [observed,name=z1] at (-3.2,1.5) {$\mathsmaller X_{i_1}$};
    \node [observed,name=z2] at (-3.2,-1.5) {$\mathsmaller X_{i_2}$};
    \node [observed,name=z3] at (1.7,1.5) {$\mathsmaller X_{i_3}$};
    \node [observed,name=z4] at (1.7,-1.5) {$\mathsmaller X_{i_4}$};
    \node [hidden,name=h] at ($(-1.5,0)$) {$\mathsmaller H$};
    \node [hidden,name=g] at ($(0,0)$) {$\mathsmaller G$};
    \node [hidden3,name=empty1] at ($(-2.3,0.7)$) {$ $};
    \node [hidden3,name=empty2] at ($(-2.3,-0.7)$) {$ $};
    \node [hidden3,name=empty3] at ($(0.8,0.7)$) {$ $};
    \node [hidden3,name=empty4] at ($(0.8,-0.7)$) {$ $};
    \draw [-] (z1) to (empty1);
    \draw [line width=0.4mm,style=dotted] (empty1) to (h);
    \draw [-] (z2) to (empty2);
    \draw [line width=0.4mm,style=dotted] (empty2) to (h);
    \draw [-] (z3) to (empty3);
    \draw [line width=0.4mm,style=dotted] (empty3) to (g);
    \draw [-] (z4) to (empty4);
    \draw [line width=0.4mm,style=dotted] (empty4) to (g);
    \draw [-] (h) to (g);
    \end{tikzpicture}
    &
    \hspace*{1cm}
    \begin{tikzpicture}
    [
      scale=0.8,
      observed/.style={circle,inner sep=0.3mm,draw=black,fill=MyBlue}, 
      hidden/.style={circle,inner sep=1.1mm,draw=black},
      hidden2/.style={circle,inner sep=0.3mm,draw=black},
      hidden3/.style={circle,inner sep=2mm,draw=black},
    ]
    \node [observed,name=z1] at (-3.7,1.5) {$\mathsmaller X_{i_1}$};
    \node [observed,name=z2] at (-3.7,-1.5) {$\mathsmaller X_{i_2}$};
    \node [observed,name=z3] at (4.2,1.5) {$\mathsmaller X_{i_3}$};
    \node [observed,name=z4] at (4.2,-1.5) {$\mathsmaller X_{i_4}$};
    \node [hidden,name=h] at ($(-2,0)$) {$\mathsmaller H$};
    \node [hidden2,name=m1] at ($(-1,0)$) {$\mathsmaller M_1$};
    \node [hidden2,name=m2] at ($(0,0)$) {$\mathsmaller M_2$};
    \node [hidden2,name=ml] at ($(1.5,0)$) {$\mathsmaller M_l$};
    \node [hidden,name=g] at ($(2.5,0)$) {$\mathsmaller G$};
    \node [hidden3,name=empty1] at ($(-2.8,0.7)$) {$ $};
    \node [hidden3,name=empty2] at ($(-2.8,-0.7)$) {$ $};
    \node [hidden3,name=empty3] at ($(3.3,0.7)$) {$ $};
    \node [hidden3,name=empty4] at ($(3.3,-0.7)$) {$ $};
    \draw [-] (z1) to (empty1);
    \draw [line width=0.4mm,style=dotted] (empty1) to (h);
    \draw [-] (z2) to (empty2);
    \draw [line width=0.4mm,style=dotted] (empty2) to (h);
    \draw [-] (z3) to (empty3);
    \draw [line width=0.4mm,style=dotted] (empty3) to (g);
    \draw [-] (z4) to (empty4);
    \draw [line width=0.4mm,style=dotted] (empty4) to (g);
    \draw [-] (h) to (m1);
    \draw [-] (m1) to (m2);
    \draw [line width=0.4mm,style=dotted]  (m2) to (ml);
    \draw [-] (ml) to (g);
    \end{tikzpicture}
    \end{tabular}
    \centering
    \caption{Topologies of quartets corresponding to a single edge $H-G$ and to a path $H-M_1-M_2-\cdots -M_l-G$.}
    \label{fig:path_topology}
\end{figure}

The recovery conditions for latent trees can be derived in two steps. 
The first step is to provide recovery conditions for those quartet relations corresponding to a single edge $H-G$ in the tree 
(Figure~\ref{fig:path_topology}, left). 
In the second step we study quartet relations corresponding to paths
$H-M_1-M_2-\cdots -M_l-G$ in the tree (Figure~\ref{fig:path_topology}, right).
We provide a condition under which the recovery condition of such quartets is reduced
to the recovery condition on quartets from step 1. That is, we provide a condition under which 
the perturbation on the path is guaranteed to be smaller than the maximum allowed perturbation on an edge.

Let
\begin{align*}
 \delta := \max_{H-G~\text{an edge}} \nbr{\Delta_{HG}}_F\,.
\end{align*}
Our goal is to obtain conditions on $\delta,$ under which recovery of any quartet relation is guaranteed.

\subsection{Quartets Corresponding to a Single Edge}
The first step is readily obtained from \S\ref{sect:perturbation} if we assume that all
CPTs (including $P_{X_{i_1}|H},\, P_{X_{i_2}|H},\, P_{X_{i_3}|G},\,P_{X_{i_4}|G}$) 
have full rank. Let $\theta_{\min} = \min_{\text{quarter}~q}{\theta_q}$. From (\ref{cond:delta_quartet}), we have
\begin{align}
\label{rel:delta_theta}
    \delta
  \leq \min \frac{\|B_\perp\|_{\ast} - \|A_\perp\|_{\ast}}{k^2+k} =\frac{\theta_{\min}}{k^2+k}.
\end{align}

\subsection{Quartets Corresponding to a Path}
{\bf Path of independent latent variables.} 
For the second step, we start again from the
fully factorized case (independent case). The joint probability table $P_{HG}$ of the two end points in a path $H-M_1-M_2-\cdots -M_l-G$ is
\begin{align*}
 P_{HG}
&= P_{H|M_1} P_{M_1|M_2} \cdots P_{M_l|G} P_G \\[1mm]
&= P_{H M_1} \diag(P_{M_1})^{-1} P_{M_1 M_2} \diag(P_{M_2})^{-1} \cdots \diag(P_{M_l})^{-1} P_{M_l G} \\[1mm]
&= P_{H} P_{M_1}^\top \diag(P_{M_1})^{-1} P_{M_1} P_{M_2}^\top \diag(P_{M_2})^{-1} \cdots \diag(P_{M_l})^{-1} P_{M_l} P_{G}^\top \\[1mm]
&= P_{H} (P_{M_1}^\top \diag(P_{M_1})^{-1}) P_{M_1} (P_{M_2}^\top \diag(P_{M_2})^{-1}) \cdots \diag(P_{M_l})^{-1} P_{M_l} P_{G}^\top \\[1mm]
&= P_{H} \one^\top P_{M_1} \one^\top \cdots \one^\top P_{M_l} P_{G}^\top \\[1mm]
&= P_{H} P_{G}^\top\,,
\end{align*}
where we have used $P_{M_i}^\top \diag(P_{M_i}(:))^{-1} = \one^\top$.

{\bf Path of dependent latent variables.}
Next, we add perturbation matrices to the joint probability tables associated with each edge $M_i-M_{j}$ in the tree and assume that the resulting joint probability table $P_{M_iM_j} = P_{M_i} P_{M_j}^\top + \Delta_{ij}$ has full rank. Furthermore, we assume that the resulting joint probability table $P_{HG}$ of the two end points in a path $H-M_1-M_2\cdots M_l-G$ also has full rank. We have
\begin{align}
 P_{HG}
&= P_{H|M_1} P_{M_1|M_2} \cdots P_{M_l|G} P_G\nonumber \\[1mm]
&= P_{H M_1} \diag(P_{M_1})^{-1} P_{M_1 M_2} \diag(P_{M_2})^{-1} \cdots \diag(P_{M_l})^{-1} P_{M_l G} \nonumber\\[1mm]
&= (P_H P_{M_1}^\top + \Delta_1) \diag(P_{M_1})^{-1} (P_{M_1} P_{M_2}^\top  + \Delta_2) \diag(P_{M_2})^{-1} \cdots \diag(P_{M_l})^{-1} (P_{M_l} P_G^\top  + \Delta_l)\nonumber\\[1mm]
&= P_{H} P_{M_1}^\top \diag(P_{M_1})^{-1} P_{M_1} P_{M_2}^\top \diag(P_{M_2})^{-1} \cdots \diag(P_{M_l})^{-1} P_{M_l} P_{G}^\top \nonumber\\[1mm]
&~~~~+ 0~~\text{(terms not involving all the $\Delta$s will all be zero)} \nonumber\\[1mm]
&~~~~+ \Delta_1 \diag(P_{M_1})^{-1} \Delta_2 \diag(P_{M_2})^{-1} \cdots \diag(P_{M_l})^{-1} \Delta_l \nonumber\\[1mm]
\label{eq:path_Delta}&= P_H P_G^\top + \Delta_1 \diag(P_{M_1})^{-1} \Delta_2 \diag(P_{M_2})^{-1}  \cdots \diag(P_{M_l})^{-1} \Delta_l\,.
\end{align}
The reason why we do not need to perturb the term $\diag(P_{M_i})^{-1}$ is that if $\widetilde P_{M_i}$ is the perturbed $P_{M_i},$
$$\widetilde P_{M_i} = \widetilde P_{M_iM_j}\, \one = (P_{M_i} P_{M_j}^\top + \Delta_{ij}) \one = P_{M_i} P_{M_J}^\top \one + \zero = P_{M_i},$$
since $\Delta_{ij}\, \one = \zero$. And the reason why terms not involving all the $\Delta$s will all be zero is that such terms contain either 
$\one^\top \Delta=\zero^\top$ or $\,\Delta\, \one = \zero$.

Now, from (\ref{eq:path_Delta}) it follows that the perturbation corresponding to the path $H-M_1-M_2-\cdots- M_l-G$ is
\begin{equation}
\Delta:=\Delta_1 \diag(P_{M_1})^{-1} \Delta_2 \diag(P_{M_2})^{-1}  \cdots \diag(P_{M_l})^{-1} \Delta_l.
\label{eq:delta_of_path}
\end{equation}

{\bf Bounding the perturbation on the path.}
We still need to show under which condition $\Delta$ from (\ref{eq:delta_of_path}) will satisfy 
$\nbr{\Delta}_F \leq \delta.$
Assume that the smallest entry in a marginal distribution of an internal node is bounded from below by $\gamma_{\min}$,~\ie,~
\begin{align*}
 \gamma_{\min} := \min_{\text{hidden node}~H} \min_{i} P_H(i)\,.
\end{align*}
Then we have
\begin{align*}
\nbr{\Delta}_F = &\nbr{\Delta_1 \diag(P_{M_1})^{-1} \Delta_2 \diag(P_{M_2})^{-1}  \cdots \Delta_l}_F \\[1mm]
 \leq &
 \nbr{\Delta_1 \diag(P_{M_1})^{-1}}_F \nbr{\Delta_2 \diag(P_{M_2})^{-1}}_F  \cdots \nbr{\Delta_l}_F \\
 \leq & \frac{\delta^l}{\gamma_{\min}^{l-1}}\,.
\end{align*}
The perturbation $\Delta$ on the path $H-M_1-M_2\cdots M_l-G$ is bounded by $\delta$ if $\frac{\delta^l}{\gamma_{\min}^{l-1}}\leq \delta$, \ie,  if 
\begin{equation}
\delta\leq\gamma_{\min}.
\label{rel:delta_gamma}
\end{equation}

From (\ref{rel:delta_theta}) and (\ref{rel:delta_gamma}) we arrive at the condition for successful quartet test for all quartets
\begin{align*}
 \delta \leq \min\cbr{\frac{\theta_{\min}}{k^2+k},\gamma_{\min}}\,.
\end{align*}
Intuitively, it means that the size of the perturbation $\delta$ away from independence can not be too large. In particular, it has to be small compared to the smallest marginal probability $\gamma_{\min}$ of a hidden state; it also has to be small compared to 
the smallest excessive dependence $\theta_{\min}$.

\section{Statistical Guarantee for the Quartet Test}
\label{app:stat:quartet}

Based on the concentration result for nuclear norm in~\eq{eq:samplebound}, we have that, given $m$ samples, the probability that the finite sample nuclear norm deviates from its true quantity by $\epsilon:=\frac{2\sqrt{2\tau}}{\sqrt{m}}$ is bounded
\begin{align}
 \PP\cbr{\|\widehat{A}\|_\ast \geq \|A\|_\ast + \epsilon} \leq 2e^{-\frac{m\epsilon^2}{8}}~~~~\text{and}~~~~
 \PP\cbr{\|\widehat{B}\|_\ast \leq \|B\|_\ast - \epsilon} \leq 2e^{-\frac{m\epsilon^2}{8}},
\end{align}
where we have used $\tau = \frac{m\epsilon^2}{8}$. Now we can derive the probability of making an error for individual quartet test. First, let 
$q=\{\{i_1,i_2\},\{i_3,i_4\}\}$ and 
\begin{align*}
 \alpha = \min \cbr{\|B(q)\|_\ast - \|A(q)\|_\ast, \|C(q)\|_\ast - \|A(q)\|_\ast}.
\end{align*}
Then, for sufficiently large $m$, we can bound the error probability by
\begin{align*}
&~~~~~~\PP\cbr{\text{Quartet test returns incorrect result}} \\
&=\PP\cbr{ \|\widehat{A}\|_\ast \geq \|\widehat{B}\|_\ast~~\text{or}~~\|\widehat{A}\|_\ast \geq \|\widehat{C}\|_\ast} \\
&\leq \PP\cbr{ \|\widehat{A}\|_\ast \geq \|\widehat{B}\|_\ast} + \PP\cbr{\|\widehat{A}\|_\ast \geq \|\widehat{C}\|_\ast}~~~~(\text{union bound})\\
&= \PP\cbr{ \|\widehat{A}\|_\ast - \|A\|_\ast + \|B\|_\ast - \|\widehat{B}\|_\ast \geq \|B\|_\ast - \|A\|_\ast } \\
&~~~~+ \PP\cbr{ \|\widehat{A}\|_\ast - \|A\|_\ast + \|C\|_\ast - \|\widehat{C}\|_\ast \geq \|C\|_\ast - \|A\|_\ast } \\
& \leq \PP\cbr{ \|\widehat{A}\|_\ast - \|A\|_\ast \geq \frac{\|B\|_\ast - \|A\|_\ast}{2}} + \PP\cbr{\|B\|_\ast - \|\widehat{B}\|_\ast \geq \frac{\|B\|_\ast - \|A\|_\ast}{2} } \\
&~~~~+ \PP\cbr{ \|\widehat{A}\|_\ast - \|A\|_\ast \geq \frac{\|C\|_\ast - \|A\|_\ast}{2}} + \PP\cbr{\|C\|_\ast - \|\widehat{C}\|_\ast \geq \frac{\|C\|_\ast - \|A\|_\ast}{2} } \\
& \leq \PP\cbr{ \|\widehat{A}\|_\ast - \|A\|_\ast \geq \frac{\alpha}{2}} + \PP\cbr{\|B\|_\ast - \|\widehat{B}\|_\ast \geq \frac{\alpha}{2} } \\
&~~~~+ \PP\cbr{ \|\widehat{A}\|_\ast - \|A\|_\ast \geq \frac{\alpha}{2}} + \PP\cbr{\|C\|_\ast - \|\widehat{C}\|_\ast \geq \frac{\alpha}{2} } \\
&\leq 8 e^{-\frac{m\alpha^2}{32}}
\end{align*}

\section{Statistical Guarantee for the Tree Building Algorithm}
\label{app:stat:tree}
Let $\alpha_q = \min \cbr{\|B(q)\|_\ast - \|A(q)\|_\ast, \|C(q)\|_\ast - \|A(q)\|_\ast}$. We define
\begin{align*}
 \alpha_{\min} = \min_{\text{quartet}~q} \alpha_q.
\end{align*}

For a latent tree with $d$ observed variables, the tree building algorithm described in the paper requires $O(d\log d)$ calls to the quartet test procedure. The probability that the tree is constructed incorrectly is bounded by the probability that either one of these quartet tests returns incorrect result. That is
\begin{align*}
&~~~~\PP\cbr{\text{The latent tree is constructed incorrectly}}  \\
&\leq \PP\cbr{\text{Either one of the $O(d\log d)$ quartet tests returns incorrect result}} \\
&\leq c \cdot d\log d \cdot \PP\cbr{\text{quartet test returns incorrect result}}~~~~(\text{union bound}) \\
&\leq 8c \cdot d\log d \cdot e^{-\frac{m\alpha^2}{32}},
\end{align*}
which implies that the probability of constructing the tree incorrectly decreases exponentially fast as we increase the number of samples $m$.

\end{document}